\documentclass{article}

\usepackage{PRIMEarxiv}

\usepackage[utf8]{inputenc} 
\usepackage[T1]{fontenc}    
\usepackage{hyperref}       
\usepackage{url}            
\usepackage{booktabs}       
\usepackage[table]{xcolor} 
\usepackage{amsfonts}       
\usepackage{nicefrac}       
\usepackage{microtype}      
\usepackage{natbib}         
\usepackage{multirow}       
\usepackage{lipsum}
\usepackage{fancyhdr}       
\usepackage{graphicx}       
\graphicspath{{images/}}     
\usepackage{graphicx}
\usepackage{amsmath} 
\usepackage{amssymb}
\usepackage{tcolorbox}
\tcbuselibrary{breakable, skins}
\usepackage{enumitem}
\usepackage{natbib} 
\usepackage{authblk}

\title{TableVerse: A Large-scale Tabletop Dataset with Real-world Grounded Layouts for Generalizable Manipulation}

\author{%
  \normalfont\normalsize
  \parbox{\textwidth}{%
    \centering
    \textbf{%
      Boyuan Wang$^{*}$, Yue Zhang$^{*}$, Xutao Xue$^{*}$,
      Xueyu Song, Yu Sun$^{\dagger}$%
    }\\
    ByteDance\\
    $^{*}$Equal contribution \qquad $^{\dagger}$Corresponding author\\
    \texttt{sun.ny@bytedance.com}%
  }%
}

\begin{document}
\maketitle

\begin{figure}[h!]
    \centering
    \vspace{-0.5cm} 
    \includegraphics[width=0.82\linewidth]{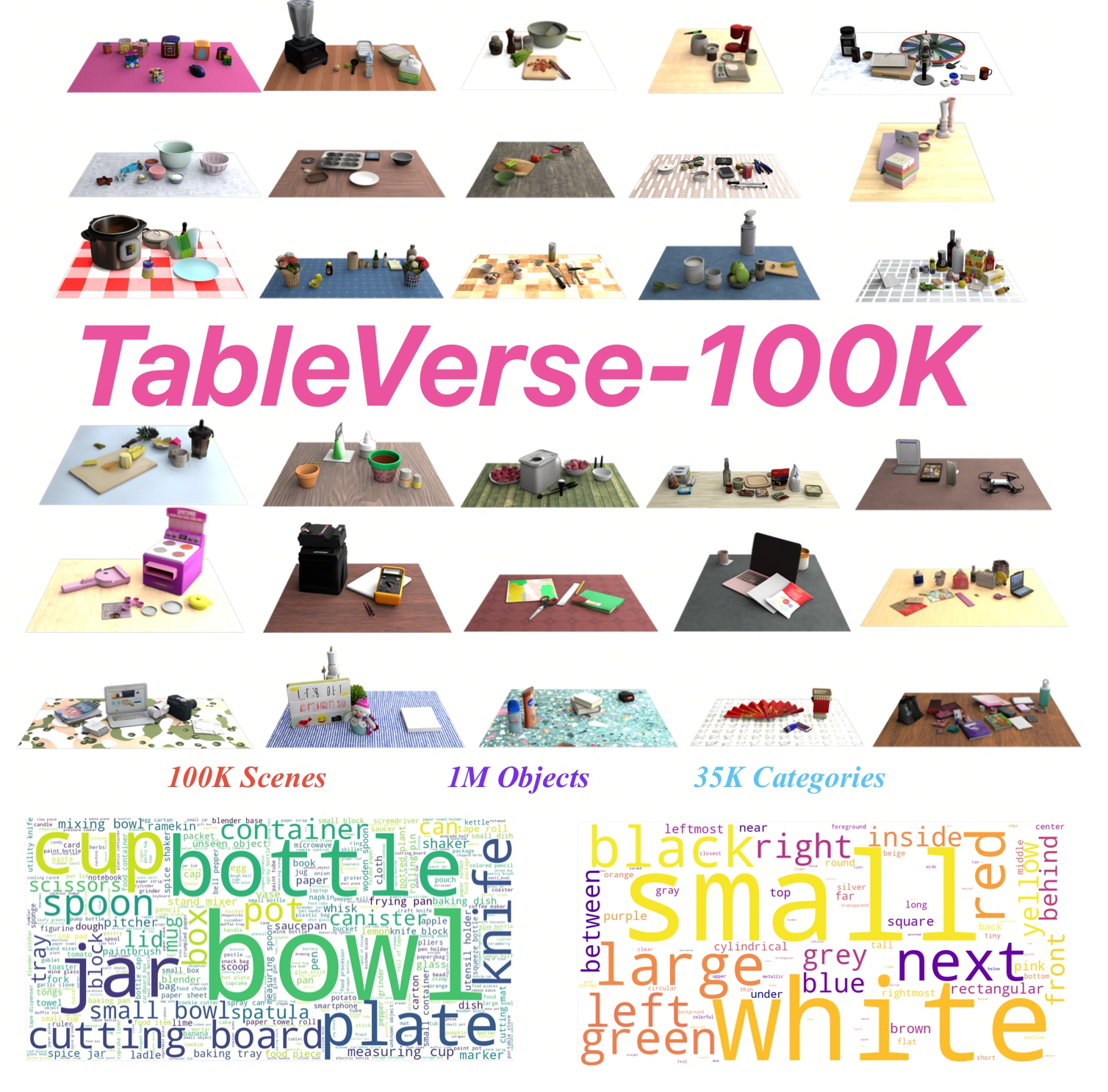}
    \vspace{-0.2cm} 
    \caption{\textbf{Statistical overview and scale of the TableVerse-100K dataset.} Our framework establishes an unprecedented milestone in automated Real2Sim tabletop asset production, encompassing \textbf{100K} physically consistent and real-world grounded interactive scenes, nearly \textbf{1M} distinct individual object instances, and spanning more than \textbf{35K} diverse semantic categories. This massive scale provides a rich, long-tail data distribution to foster robust, highly generalizable visuomotor policy learning for robotic manipulation.}
    \label{fig:teaser_dataset_stats}
\end{figure}

\begin{abstract}
The development of generalizable robotic manipulation policies is inherently bounded by the availability of large-scale, high-fidelity scene data. 
While recent automated synthesis methods attempt to bridge this gap via text-to-layout hallucination or simplified procedural generation, they frequently suffer from physical implausibility and fail to capture the complex, dense clutter of actual human environments. 
In this paper, we introduce \textbf{TableVerse}, a fully automated Real2Sim pipeline that shifts the paradigm from imaginative layout generation to deterministic reconstruction from unstructured, \textit{in-the-wild} image data. 
Our framework seamlessly processes unscripted internet media into high-fidelity, simulation-ready tabletop environments with accurate metric scales, authentic topologies, and verified mechanical stability. 
Furthermore, an automated task-conditioned trajectory generation framework is integrated to synthesize high-quality, collision-free pick-and-place demonstrations. 
Leveraging this complete pipeline, we construct the \textbf{TableVerse-100K Dataset}, a large-scale corpus comprising 100,000 unique, physically consistent environments paired with interactive manipulation trajectories. 
By capturing diverse asset compositions, realistic spatial distributions, and high-quality demonstrations, TableVerse-100K establishes a highly scalable and high-fidelity data foundation, providing significant value to facilitate future research in generalizable robotic manipulation tasks.
Our project page is available at \url{https://bytedance.github.io/TableVerse}.
\end{abstract}


\section{Introduction}
Generalizable robotic manipulation demands large-scale, physics-ready simulation data that faithfully mirrors real-world layout distributions. However, existing automated generation methods are typically constrained to synthesizing simplistic, sparse layouts and suffer from severe geometric collisions that render them unusable for stable physics simulation. Confronted with unstructured, \textit{in-the-wild} real-world data, these conventional paradigms fail catastrophically, proving completely incapable of reconstructing the dense clutter and complex topologies characteristic of actual human environments.

To bridge this gap, we introduce \textbf{TableVerse}, a scalable real-to-sim pipeline that converts unstructured internet media directly into interactive simulation environments. Our framework orchestrates a seamless perception-to-simulation workflow: it extracts and deconstructs tabletop assets from single-view observations to restore accurate metric scales, applies a layout-preserving geometric optimization to disentangle intersecting meshes, and utilizes MuJoCo physics stabilization to settle objects into realistic resting states. Finally, TableVerse deploys accelerated motion planning to automatically synthesize task-conditioned, collision-free expert trajectories, transforming raw web data into fully interactive digital twins.

Leveraging this closed-loop workflow, we construct the \textbf{TableVerse-100K Dataset}, comprising $100,000$ unique, physically consistent tabletop environments paired with continuous expert manipulation trajectories. To our knowledge, this represents one of the largest and most physically faithful datasets for tabletop manipulation, capturing the dense clutter and heterogeneous physics of real human environments to provide a highly scalable data foundation for scaling up downstream policy learning.

In summary, our main contributions are three-fold. First, we develop the \textbf{TableVerse Pipeline}, an automated, observation-driven Real2Sim framework that seamlessly converts unscripted, \textit{in-the-wild} internet media into high-fidelity, simulation-ready environments. Second, we introduce the \textbf{LCCR Optimization} framework, a layout-consistent geometric rectification module designed to elegantly disentangle intersecting meshes via hierarchical scene graphs, which is tightly coupled with a physics-based resting-state stabilization step. Third, we construct and release the \textbf{TableVerse-100K Dataset}, a massive data foundation consisting of $100,000$ diverse, physics-grounded tabletop scenes augmented with continuous expert trajectories, thereby providing a highly scalable benchmark to facilitate future research in generalizable robotic manipulation.

\section{Related Work}

\subsection{3D Scene Reconstruction and Layout Synthesis}
Automated 3D tabletop generation spans vision-driven reconstruction and language-conditioned layout synthesis. Within the visual domain, single-view methods fail to balance geometric fidelity with physical feasibility: MIDI~\citep{huang2025midi} degrades under dense occlusion despite multi-instance attention; SAM3D~\citep{chen2025sam} regresses 3D poses via layout tokens and point maps but lacks metric constraints; and SceneMaker~\citep{shi2025scenemaker} induces severe mesh interpenetrations in dense clusters. Parallelly, language-conditioned paradigms deploy LLMs to synthesize scenes via text queries~\citep{Feng2023LayoutGPTCV, Yang2023HolodeckLG}, symbolic graphs~\citep{hao2026mesatask}, or generated 2D priors~\citep{wang2025tabletopgen}. However, lacking continuous geometric grounding, these approaches suffer from severe scale errors. Crucially, bounded by abstract symbolic priors, LLM layouts remain overly simplistic and scattered, failing to capture the dense clutter, vertical stacking, and composite asset structures of real-world environments, thereby leaving a massive reality gap for robot manipulation.

\subsection{Tabletop Scene Datasets}
While large-scale indoor room-level datasets are abundant~\citep{yu2025metascenes, Fu20203DFRONT3F, Dai2017ScanNetR3, Zhou2025IL3DAL}, specialized benchmarks focused exclusively on tabletop environments remain limited. Existing pipelines rely heavily on synthetic asset composition: TO-Scene~\citep{Xu2022TOSceneAL} populates simulation scenes with CAD models via a simplistic top-down ``click-and-drop'' mechanism, which completely neglects complex hierarchical nesting or vertical stacking. Alternatively, MesaTask-10K~\citep{hao2026mesatask} uses text-to-layout generation and asset retrieval. However, it suffers from an over-idealized neatness bias, producing sparse and orderly layouts that fail to mirror the dense, chaotic, and cluttered distributions of real-world human environments. By shifting the paradigm from synthetic imagination to deterministic \textit{in-the-wild} reconstruction, our TableVerse-100K dataset bridges these gaps, delivering unprecedented scale alongside physically grounded, authentic desktop distributions.

\subsection{6-DoF Grasp Synthesis and Robotic Motion Planning}
Efficient demonstration generation relies on robust 6-DoF grasp synthesis and collision-free motion planning. While early analytical or data-driven grasping methods heavily depend on pre-modeled templates or heuristic descriptors~\citep{mahler2017dex, ten2017grasp, liang2019pointnetgpd}, modern dense architectures directly map raw point clouds to continuous 6-DoF grasp spaces~\citep{sundermeyer2021contact, fang2020graspnet}. To handle severe visual occlusions in dense clutter, GraspGen~\citep{murali2025graspgen} introduces a generative diffusion framework for robust pose sampling, which we adopt to synthesize high-quality candidate grasps. For trajectory execution, conventional optimization planners and foundation-model-driven controllers have been widely explored~\citep{schulman2013finding, liang2023code, huang2023voxposer}, culminating in LLM-driven multi-stage simulation workflows like GenManip~\citep{gao2025genmanip}. However, these synthesis pipelines remain bottlenecked by text-to-layout spatial hallucinations or over-simplified object arrangements. In contrast, TableVerse pairs observation-driven, chaotic Real2Sim environments with parallelized, GPU-accelerated motion optimization via cuRobo~\citep{sundaralingam2023curobo}, successfully generating a massive, physically validated expert trajectory corpus.

\begin{figure}[ht]
    \centering                                          
    \includegraphics[width=1.0\linewidth]{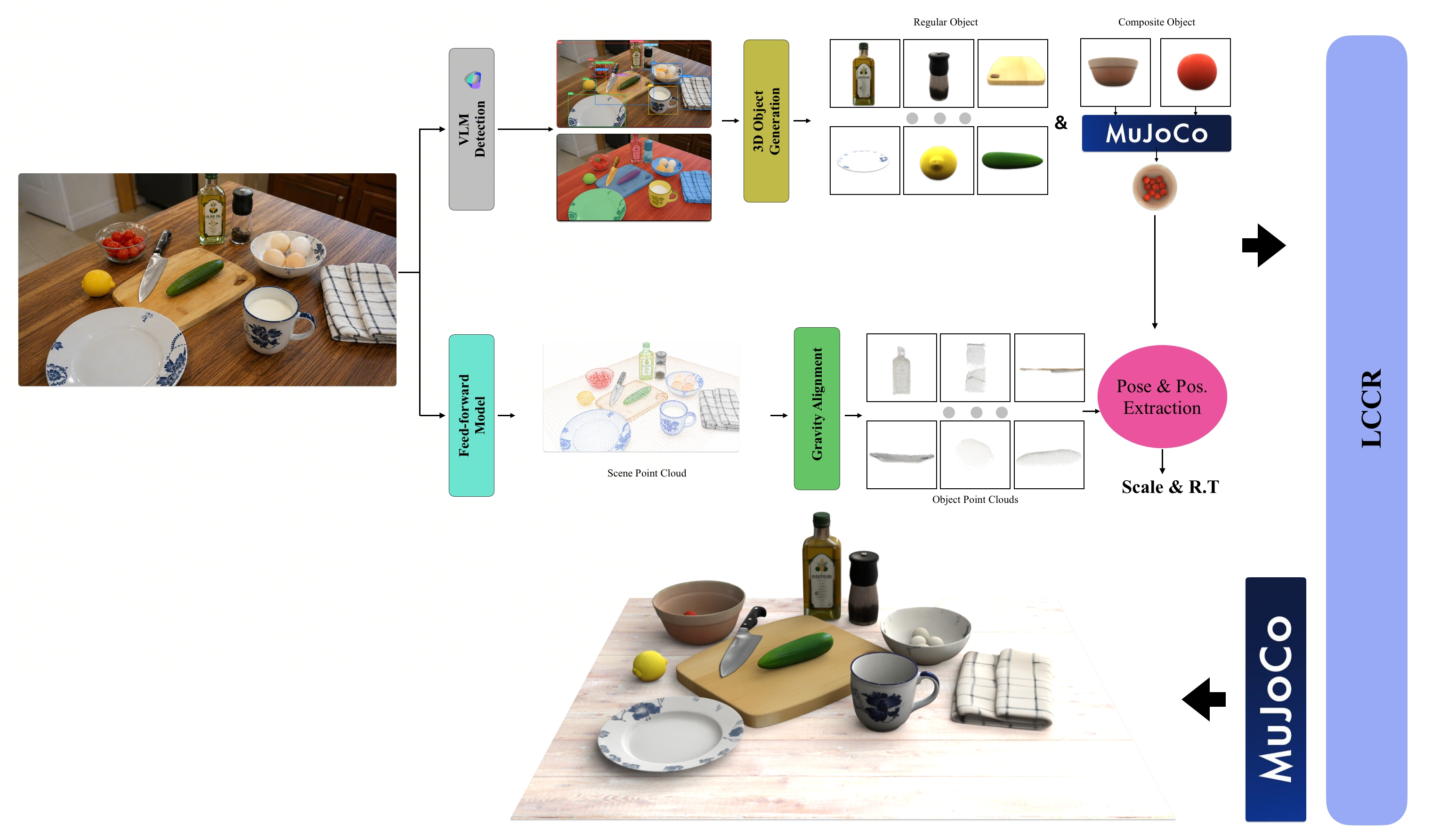}
    \caption{\textbf{Overview of the TableVerse pipeline for automated tabletop scene synthesis.} Given an unstructured, single-view real-world observation, our framework constructs physics-ready digital twins through four sequential stages: (1) \textbf{Instance Extraction:} performing open-vocabulary object detection and high-fidelity mask generation; (2) \textbf{Composite Asset Deconstruction:} uncoupling nested entities via 3D reconstruction and isolated free-fall assembly; (3) \textbf{Metric Scale \& Pose Alignment:} extracting scene point clouds and executing a coarse-to-fine registration to regress precise dimensions and 6-DoF poses; and (4) \textbf{LCCR \& Physics Stabilization:} resolving mesh interpenetrations via a layout-consistent geometric optimization under topological constraints, followed by a final MuJoCo simulation to settle objects into a mechanically stable resting state.}
    \label{fig:figure1}
\end{figure}

\section{Method}
In this section, we present our novel monocular real-to-sim framework \textbf{TableVerse} for generating high-fidelity, simulation-ready 3D tabletop environments and continuous expert trajectories directly from unstructured internet media. The crux of our approach lies in replacing probabilistic spatial hallucinations with a deterministic perception-to-physics workflow, thereby endowing the pipeline with the capability to preserve authentic real-world metric scales, object topologies, and contact mechanics. 

We outline the comprehensive system workflow in Figure~\ref{fig:figure1}. Specifically, we elaborate on our open-vocabulary hybrid object extraction, structural composite asset deconstruction, and coarse-to-fine 6-DoF pose registration in Section 3.1. To resolve initial geometric alignment noise and completely eliminate mesh interpenetrations while strictly safeguarding the macroscopic spatial layout, we introduce our novel \textbf{Layout-Consistent Collision Rectification (LCCR)} module in Section 3.2. Leveraging this closed-loop automated pipeline, we detail the scale-up synthesis, MLLM-driven validation, and curation of the massive \textbf{TableVerse-100K Dataset} in Section 3.3. Finally, Section 3.4 presents our task-conditioned trajectory generation framework, which translates high-level manipulation specifications into continuous, collision-free robotic joint-space paths within the reconstructed digital twins.

\begin{figure}[htbp]
    \centering
    \includegraphics[width=0.7\linewidth]{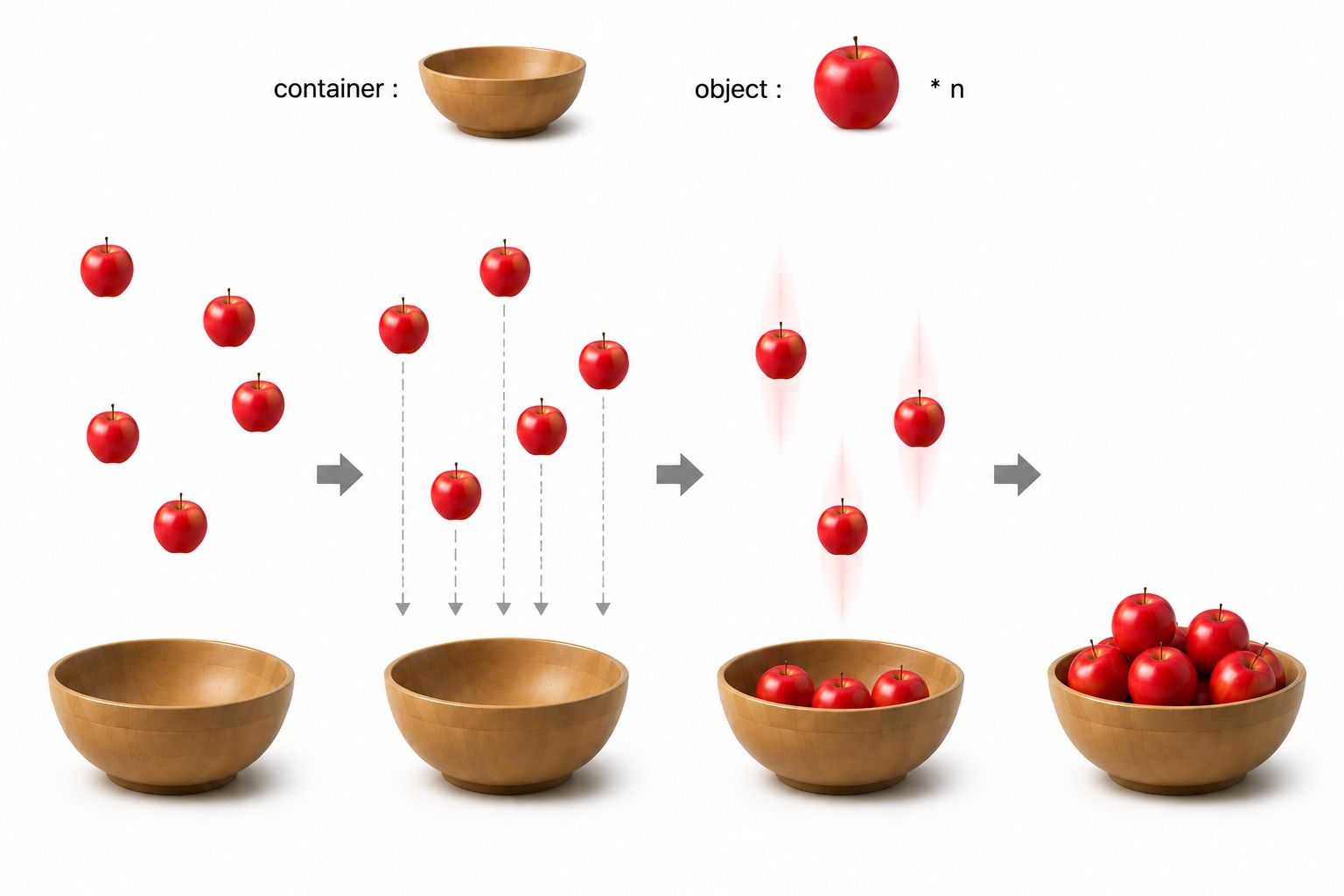}
    \caption{\textbf{Physics-guided assembly workflow for composite object generation.} To enrich asset diversity and interaction complexity during the manipulation process, our pipeline structurally uncouples composite entities. Individual components (the container and its nested contents) are reconstructed as separate isolated meshes via SAM3D, and subsequently initialized within an isolated MuJoCo environment. Running a brief free-fall simulation allows the internal items to naturally ``drop'' and settle into the container, establishing valid physical contacts and preserving independent structural manipulation properties for downstream tasks.}
    \label{fig:composite_assembly}
\end{figure}

\subsection{Instance-level Object Extraction and Generation}

\paragraph{Open-Vocabulary Hybrid Object Detection} 
To bypass the error accumulation of cascading an MLLM with GroundingSAM-v2~\citep{ren2024grounded}, we employ Seed-1.8~\citep{Seed2026Seed18MC} for direct open-vocabulary detection on the tabletop scene. Assets are classified into \textit{regular objects} and \textit{composite objects} (containers holding nested contents), with non-meshable substances (e.g., liquids, powders) falling back to regular single entities. For interior item clusters, the detector outputs a single representative bounding box alongside an instance count to avoid spatial ambiguity. These bounding boxes are subsequently forwarded to SAM2~\citep{ravi2025sam} to extract precise instance segmentation masks.

\paragraph{Composite Object Generation} 
To mitigate severe occlusions common in \textit{in-the-wild} images, we utilize SAM3D to reconstruct 3D meshes from the extracted masks. For composite objects, the container and its nested contents are reconstructed as separate individual meshes. As illustrated in Figure~\ref{fig:composite_assembly}, we then initialize these assets in an isolated MuJoCo~\citep{Todorov2012MuJoCoAP} free-fall simulation, ``dropping'' the contents into the container. This physics-guided assembly yields a physically valid composite asset where all constituent objects maintain independent structural manipulation properties.

\paragraph{Object Position and Pose Extraction} 
To recover accurate metric scales and spatial intervals, we leverage Depth Anything 3~\citep{lin2025depth} to extract scene point clouds from the monocular input. Crucially, prior to segment decomposition, we utilize the segmented table or floor surface mask to estimate the dominant plane normal and establish the absolute gravity vector. A global gravity-alignment transformation is subsequently applied to rectify the scene orientation into a canonical coordinate system, ensuring that the downward physics vectors align seamlessly with simulation realities. Following this coordinate normalization, object-specific point cloud segments are structurally isolated via the SAM2 masks. To eliminate capture-induced sensing artifacts and background interference, each isolated segment undergoes statistical and radius-based denoising to explicitly purge stray noise and sporadic outlier points. We then perform a coarse-to-fine alignment between the cleaned, observed point cloud and the sampled mesh vertices. Specifically, we execute a continuous search over the $z$-axis rotation to identify the optimal initial heading with the minimal Chamfer distance, followed by Iterative Closest Point (ICP) registration to accurately regress the final 6-DoF poses and dimensions.

\subsection{Layout-Consistent Collision Rectification} 
Directly deploying initial layouts from raw registration often yields severe mesh interpenetration due to asset discrepancies and optimization noise. Naively resolving these errors by directly initializing the physics engine generates massive repulsive torques that destabilize the scene, a failure mode exacerbated by dense, \textit{in-the-wild} clutter. Critically, while standard physics simulators can inherently resolve minor, microscopic interpenetrations through numerical contact relaxation, they fail catastrophically when confronted with deep geometric intersections. Instead of settling smoothly, deeply intersecting geometries trigger explosive, non-physical constraint forces that instantly scatter or permanently distort the scene clutter. It is therefore imperative to explicitly pre-rectify these severe macroscopic overlaps through geometric optimization before delegating the layout to the simulation engine for final mechanical stabilization. 

\begin{figure}[ht]
    \centering 
    \includegraphics[width=1.0\linewidth]{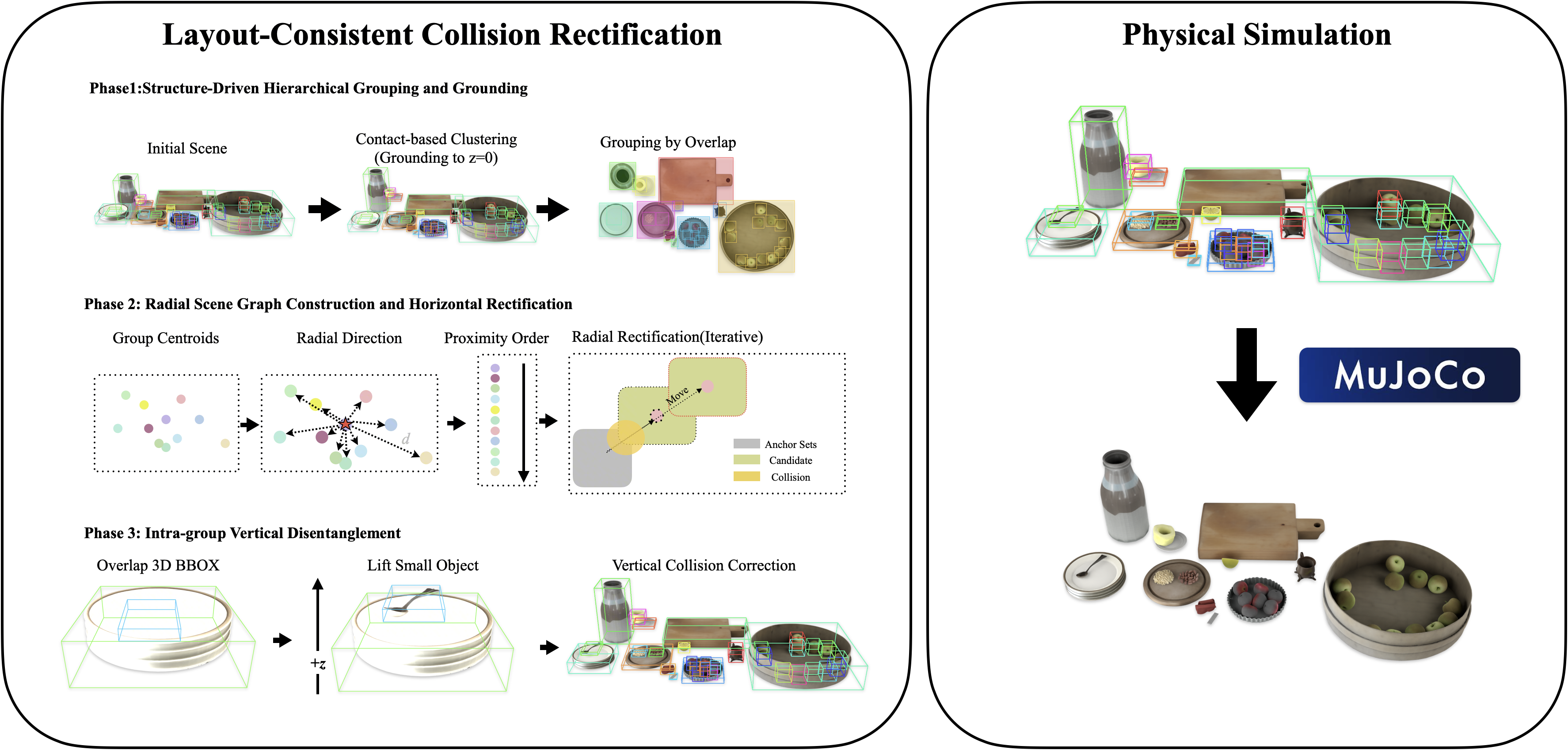}
    \caption{\textbf{Detailed schematic of the Layout-Consistent Collision Rectification (LCCR) module.} The pipeline ingests initially registered overlapping meshes and resolves spatial intersections through three sequential phases: (1) \textbf{Hierarchical Contact Grouping \& Grounding:} clustering adjacent assets and decoupling intentional vertical stacking from horizontal clutter via a 2D overlap threshold; (2) \textbf{Radial Graph Construction \& Horizontal Rectification:} building a proximity-driven radial scene graph and rigidly translating entities along azimuthal layout vectors to eliminate lateral overlap; and (3) \textbf{Vertical Disentanglement \& Physics Stabilization:} adjusting z-axis heights within stacked groups to resolve remaining vertical intersections, followed by a gravity-driven MuJoCo simulation to naturally settle objects into a mechanically stable resting state.}
    \label{fig:figure2}
\end{figure}

To address this, we introduce the \textbf{Layout-Consistent Collision Rectification (LCCR)} module, whose detailed operational schematic is illustrated in Figure~\ref{fig:figure2}. LCCR elegantly disentangles intersecting meshes while strictly preserving the macroscopic spatial layout through a sequential three-phase workflow, as detailed below.

\paragraph{Phase 1: Hierarchical Contact Grouping and Grounding}
We first cluster objects in physical contact (treating composite assets as single entities) and ground all groups by translating their lowest vertices to the table plane ($z=0$). To isolate vertical stacking from horizontal registration noise, we project all assets onto the horizontal plane to extract top-view 2D bounding boxes. Objects exhibiting a significant area overlap ratio ($\ge 50\%$) are merged into a unified hierarchical group to safeguard containment or stacking relations. Conversely, pairs with $< 50\%$ overlap are treated as disjoint horizontal entities, effectively decoupling accidental lateral clipping from intentional vertical arrangements.

\paragraph{Phase 2: Radial Graph Construction and Horizontal Rectification}
To resolve horizontal collisions layout-preservingly, we construct a proximity-driven radial scene graph that governs an incremental expansion workflow. Let $\mathcal{G} = \{G_1, G_2, \dots, G_N\}$ denote the set of groups with 2D bounding box centroids $c_i \in \mathbb{R}^2$. The most spatially central group is uniquely defined as the root $G_{\text{root}}$ at the coordinate origin. For any other group $G_i$, its relative horizontal layout is parameterized by a radial directional unit vector $\vec{v}_i$:
\begin{equation}
\vec{v}_i = \frac{c_i - c_{\text{root}}}{\|c_i - c_{\text{root}}\|_2}
\end{equation}

To eliminate lateral overlap without altering the macroscopic azimuthal arrangement, the rectification proceeds in an inside-out, sequential manner. We first sort all non-root groups $\mathcal{G} \setminus \{G_{\text{root}}\}$ in ascending order based on their initial Euclidean distance to the root, forming an ordered sequence $\mathcal{S} = (G_{(1)}, G_{(2)}, \dots, G_{(N-1)})$, where $\|c_{(1)} - c_{\text{root}}\|_2 \le \|c_{(2)} - c_{\text{root}}\|_2 \le \dots \le \|c_{(N-1)} - c_{\text{root}}\|_2$. We then initialize an active anchor set containing only the fully settled meshes, starting with $\mathcal{A}_0 = G_{\text{root}}$. For each step $k = 1, 2, \dots, N-1$ in the sorted sequence, the optimal translation distance $d_{(k)}^*$ for the $k$-th group $G_{(k)}$ is determined by searching outward along its specific radial heading $\vec{v}_{(k)}$ until it clears all previously rectified boundaries:
\begin{equation}
d_{(k)}^* = \min \{ d \ge 0 \mid (G_{(k)} + d \vec{v}_{(k)}) \cap \mathcal{A}_{k-1} = \emptyset \}
\end{equation}
Upon solving Eq. (2), $G_{(k)}$ is rigidly translated to its layout-preserved, collision-free coordinate $c_{(k)}^* = c_{(k)} + d_{(k)}^* \vec{v}_{(k)}$, and the active anchor set is recursively augmented to incorporate the newly settled entity:
\begin{equation}
\mathcal{A}_k = \mathcal{A}_{k-1} \cup \{ G_{(k)} + d_{(k)}^* \vec{v}_{(k)} \}
\end{equation}
By iterating through the graph sequence $\mathcal{S}$, this expanding optimization guarantees the elimination of lateral overlap while strictly preserving the original azimuthal angular semantics.

\paragraph{Phase 3: Vertical Disentanglement and Physics Stabilization}
Finally, we resolve internal vertical interpenetrations within the high-overlap groups identified in Phase 1. If a 3D intersection is detected within a group, the object with the smaller horizontal footprint (projected $xy$-area) is translated upward along the positive $z$-axis until it is strictly disjoint from the supporting mesh underneath. 

Following this geometric rectification, the entirely collision-free layout is imported into the MuJoCo\citep{Todorov2012MuJoCoAP} physics engine. Because the assets enter the engine in a geometrically decoupled state, they completely bypass explosive contact forces. A brief forward simulation under standard gravity allows the objects to settle naturally, closing micro-gaps and establishing stable mechanical contact dynamics for downstream tasks.

\subsection{TableVerse-100K Dataset}

\begin{figure}[htbp]
    \centering
    \includegraphics[width=1.0\linewidth]{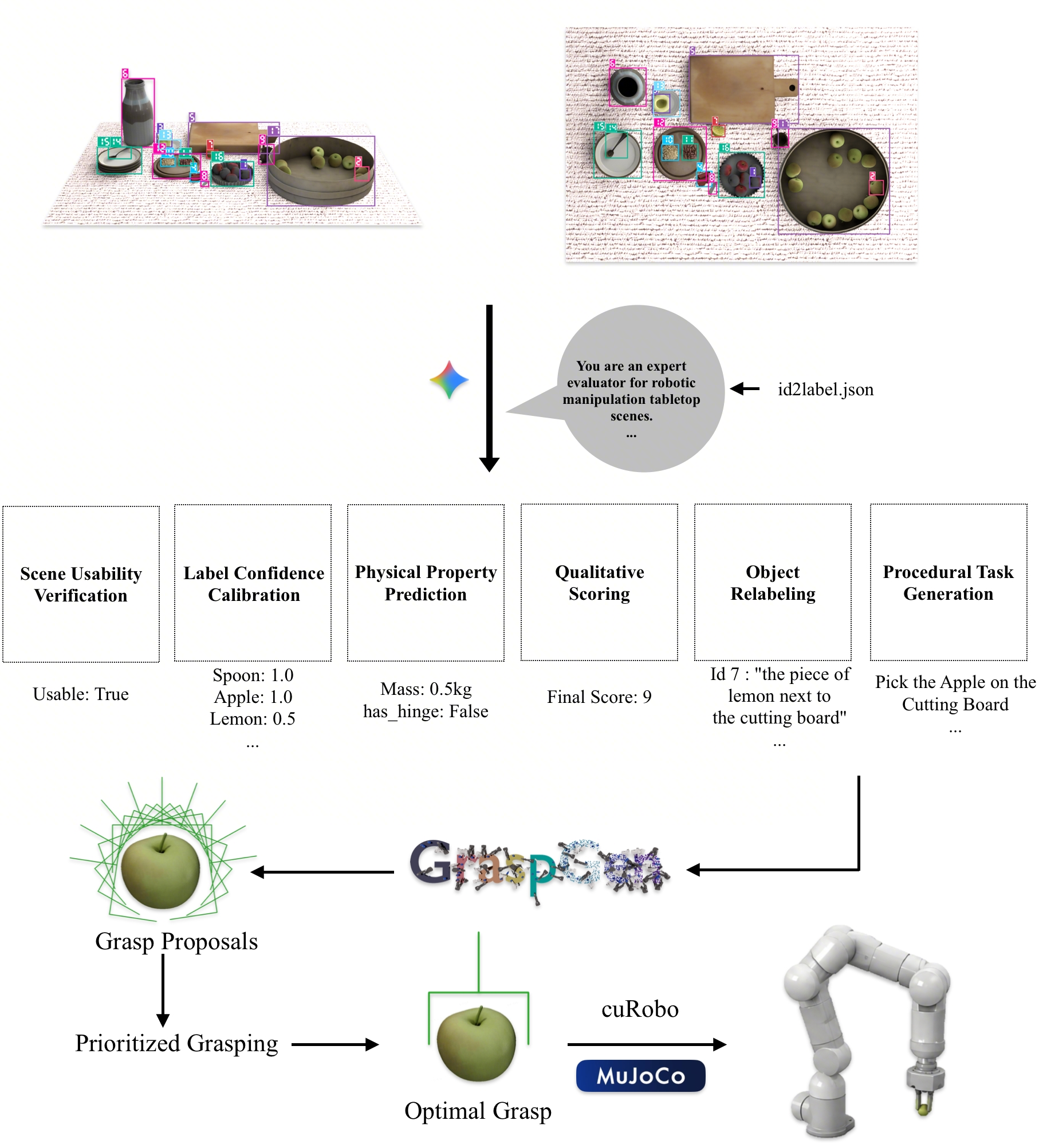}
    \caption{\textbf{Detailed schematic of the MLLM-driven curation and annotation pipeline.} Multi-view orthographic and bird's-eye renderings are concatenated into a visual composite for Gemini 2.5 Pro to execute a coordinated seven-dimensional scenario assessment, physical attribute prediction, and structured pick-and-place task generation.}
    \label{fig:app_mllm_pipeline}
\end{figure}

Leveraging our automated Real2Sim pipeline, we scale up the synthesis workflow to construct the \textbf{TableVerse-100K Dataset}, comprising 100K unique, physically consistent, and instruction-annotated tabletop environments. In total, this unprecedented scale encompasses nearly 1M distinct individual object instances spanning more than 35K diverse semantic categories, establishing a highly varied long-tail distribution for generalizable policy learning. The data sourcing pipeline targets unstructured internet images, executing automated visual filtering to explicitly isolate in-the-wild snapshots containing valid tabletop surfaces. These chaotic real-world images are subsequently ingested into our reconstruction workflow to yield an initial pool of simulated 3D digital twins.

\begin{figure}[htbp]
    \centering
    \includegraphics[width=1.0\linewidth]{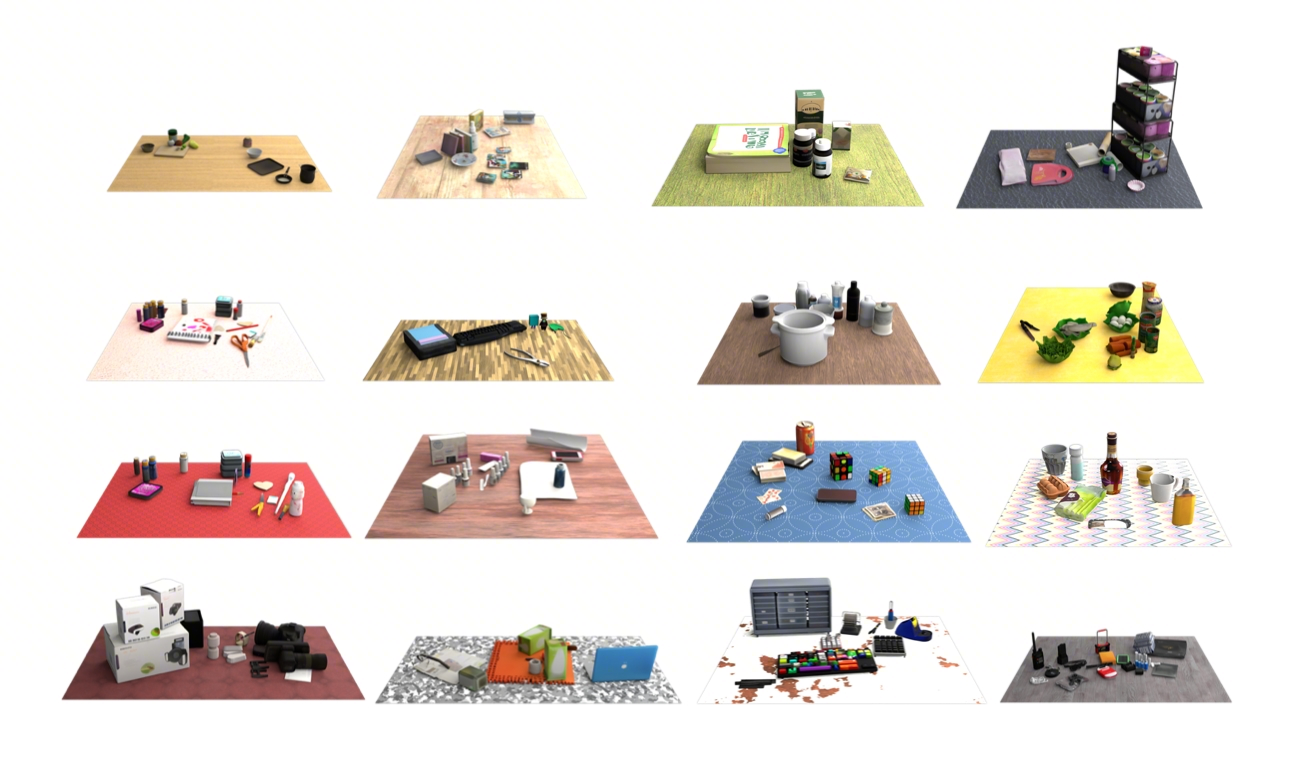}
    \caption{\textbf{Visualization of diverse tabletop texture augmentations in TableVerse.} Over 1,700 unique, high-resolution texture maps are dynamically applied to the table surfaces to expand visual diversity for robust domain randomization.}
    \label{fig:tabletop_textures}
\end{figure}

To filter out degenerate layouts and enrich the environments with high-quality metadata, we implement an automated curation and annotation workflow driven by Gemini 2.5 Pro~\citep{google2025gemini25}, as outlined in Figure~\ref{fig:app_mllm_pipeline}. For each candidate scene, we procedurally render a horizontally concatenated high-resolution composite image pairing a \textit{Front View} (capturing vertical layering) with a \textit{Top-Down View} (exposing horizontal layout). Given this multi-view visual prompt and user-provided label mappings, the MLLM acts as a rigorous data oracle to execute a deterministic seven-fold evaluation and annotation pipeline:
\begin{enumerate}[label=(\Alph*), leftmargin=2.0em, topsep=0.3ex, itemsep=0.3ex]
    \item \textbf{Usability Gate:} Executes a strict binary plausibility filter. A scene is invalidated if it exhibits extreme regional partitions (isolated clusters violating single robot workspace coverage) or non-tabletop entities (e.g., humans or large appliances), capping the final score at 4.
    \item \textbf{Severe Label Error Check:} Implements a high-threshold semantic sanity check. It filters out glaring perception errors while explicitly tolerating same-category synonyms, hypernyms, low-poly geometries, or brand-level visual inaccuracies to ensure robust data filtering.
    \item \textbf{Label Confidence Calibration:} Assigns a continuous alignment confidence score in $[0, 1]$ for every instance, where scores below 0.7 enforce conservative property predictions.
    \item \textbf{Physical Property Prediction:} Infers intrinsic material and structural dynamics for each asset, predicting its absolute mass in kilograms alongside a kinematic audit to determine whether the object contains an articulated hinge structure.
    \item \textbf{Quality Scoring:} Assigns a synthesized score from 1 to 10 governed by a multi-criteria weighted schema: object count and category diversity (45\%, favoring $n \ge 5$), geometric plausibility (15\%), robot graspability (15\%), and layout clarity (25\%).
    \item \textbf{Object Relabeling:} Completely independent of original annotations, it assigns a visual-only category name alongside an array of view-invariant, context-grounded distinguishing descriptive phrases, strictly forbidding artificial instance indices.
    \item \textbf{Pick \& Place Task Generation:} Procedurally synthesizes a comprehensive list of pick-and-place manipulation commands mapped to 10 distinct spatial relation enums under strict physical clearance, daily-life plausibility, and initial-state constraints.
\end{enumerate}

Ultimately, all verified layouts, shapes, and articulated topologies are exported as fully compliant, simulation-ready \textbf{MJCF (MuJoCo XML)} files, encapsulating absolute metric transformations and predicted contact dynamics. Furthermore, to maximize visual diversity and prevent downstream visual-motor policies from overfitting to homogeneous backgrounds, we implement large-scale domain randomization. We curate a massive repository of over 1,700 high-resolution texture maps aggregated from professional texture platforms (including Sharetextures\footnote{\url{https://www.sharetextures.com/}}, AmbientCG\footnote{\url{https://ambientcg.com/}}, and CC0-Textures\footnote{\url{https://cc0-textures.com/}}), which are dynamically mapped onto the tabletop surfaces during simulation initialization (Figure~\ref{fig:tabletop_textures}). This extensive appearance-level augmentation significantly reinforces the robustness and occlusion-resistance of downstream policy learning.

\subsection{Task-Conditioned Trajectory Generation}

Based on the collision-free reconstructed scene layouts and task specifications, we propose an automated task-conditioned trajectory generation framework to translate high-level manipulation instructions into continuous, executable robotic trajectories. As illustrated in Figure~\ref{fig:app_mllm_pipeline}, the framework establishes a closed-loop coupling between the MLLM-driven scenario assessment and downstream trajectory synthesis, routing the procedurally generated task instructions directly into a specialized manipulation loop comprising prioritized 6D grasp selection, relation-constrained placement sampling, and physics-validated motion execution.

\paragraph{Top-Down Prioritized 6D Grasp Synthesis}
To generate high-quality grasps, the target object's surface geometry is first processed by GraspGen~\citep{murali2025graspgen} to predict an array of candidate 6D grasp poses $\mathcal{G} = \{g_i\}_{i=1}^M$. Each candidate $g_i = (\mathbf{R}_i, \mathbf{t}_i)$ is parameterized by a rotation matrix $\mathbf{R}_i \in SO(3)$ and a translation vector $\mathbf{t}_i \in \mathbb{R}^3$. Let $\vec{z}_{\text{local}} = [0, 0, 1]^T$ denote the canonical tool approach axis defined in the local gripper frame. The grasp approach vector transformed into the world frame is thus expressed as $\vec{a}_i = \mathbf{R}_i \vec{z}_{\text{local}}$. To guarantee grasp stability during transit, we implement a top-down prioritization filter based on the directional alignment between $\vec{a}_i$ and the world vertical axis $\vec{z}_w = [0, 0, 1]^T$. A grasp candidate is selected as a high-quality pose if it satisfies the strict directional constraint:
\begin{equation}
\vec{a}_i \cdot \vec{z}_w \le \gamma_{\text{strict}}
\end{equation}
where $\gamma_{\text{strict}}$ is a negative threshold enforcing a near-vertical downward approach angle. Candidates failing this criterion are adaptively relaxed or discarded to maximize dynamic contact reliability.

\paragraph{Relation-Constrained Placement and Motion Planning}
To fulfill spatial placement predicates $\mathcal{R}$ (e.g., $\text{top}$, $\text{in}$), target placement poses $\mathbf{T}_{\text{place}}$ are sampled within a localized bounding region derived from the reference asset's geometry. Candidate poses are verified via an Axis-Aligned Bounding Box module to ensure sufficient clearance against all surrounding assets $\mathcal{O}_{\text{adj}}$:
\begin{equation}
|\Delta x| \ge b_{\text{src},x} + b_{\text{adj},x} + \epsilon \quad \land \quad |\Delta y| \ge b_{\text{src},y} + b_{\text{adj},y} + \epsilon
\end{equation}
where $\Delta x$ and $\Delta y$ represent the relative horizontal distances between centroids, $b_{\cdot, x}$ and $b_{\cdot, y}$ denote bounding box half-extents, and $\epsilon$ represents a physical clearance safety margin. Following the structured six-phase manipulation strategy outlined in GenManip~\citep{gao2025genmanip} (comprising pre-grasp, grasp, post-grasp, pre-place, place, and post-place phases), collision-free joint-space trajectories connecting the sampled poses are optimized and executed using the GPU-accelerated cuRobo~\citep{sundaralingam2023curobo} motion planner.

\section{Experimental Results}
\label{sec:result}

\subsection{Setup}

\paragraph{Implementation Details}
Our pipeline integrates specialized modules for open-world perception, geometry processing, and verification. We employ \textbf{Seed-1.8}~\citep{Seed2026Seed18MC} for open-vocabulary object detection, forwarding predicted bounding boxes to \textbf{SAM2}~\citep{ravi2025sam} for high-fidelity instance segmentation, where the extracted object masks are subsequently ingested by \textbf{SAM3D} to reconstruct initial 3D meshes for each individual asset. Concurrently, \textbf{Depth Anything 3}~\citep{lin2025depth} extracts metric scene point clouds from single-view inputs. To establish a physically consistent coordinate system, we leverage the segmented table or floor surface mask to compute a gravity-alignment transformation, explicitly rectifying the global orientation of the extracted point cloud prior to mesh registration. To prevent MuJoCo simulation artifacts, all 3D object meshes undergo approximate convex decomposition via \textbf{CoACD}~\citep{wei2022approximate}. Finally, we render orthogonal multi-view projections for each scene and utilize \textbf{Gemini 2.5 Pro}~\citep{google2025gemini25} for multi-dimensional scene evaluation, confidence filtering, and task instruction annotation.

\paragraph{Baselines}
We evaluate TableVerse against three representative single-view 3D scene reconstruction methods: \textbf{MIDI}~\citep{huang2025midi}, \textbf{SAM3D}~\citep{chen2025sam}, and \textbf{SceneMaker}~\citep{shi2025scenemaker}. To isolate 3D layout evaluation from upstream perception failures and ensure a fair comparison, we provide all baselines with the exact instance masks generated by our pipeline. Crucially, as these baselines cannot model hierarchical composite assets, we only provide them with the mask of the outermost container, omitting nested interior objects to align with their monolithic asset assumptions.

\paragraph{Evaluation Metrics}
To demonstrate our framework's capability in producing simulation-ready scenes with accurate metric scales, we evaluate environments across two primary dimensions:
(1) \textbf{Scene Collision Rate (\%)} $\downarrow$: The percentage of generated scenes exhibiting volumetric mesh interpenetration (clipping) prior to physics relaxation, reflecting initial geometric validity.
(2) \textbf{GPT-Score}: A multi-dimensional MLLM evaluation encompassing three sub-dimensions scored from 1 to 10---\textit{Layout Fidelity (LF)} $\uparrow$ (spatial layout and scale alignment), \textit{Visual Quality (VQ)} $\uparrow$ (texture and rendering naturalness), and \textit{Geometry Quality (GQ)} $\uparrow$ (preservation of correct categories, shapes, and structural details without distortion or collapse)---alongside the \textit{Average Rank (Avg. Rank)} $\downarrow$ which measures the mean cross-method preference ranking across all test scenes.

\subsection{Comparisons with Alternative Methods}
To evaluate the generation quality and physical fidelity of our pipeline, we construct a test set containing $100$ unscripted tabletop samples curated from raw internet images. This benchmark covers diverse real-world corner cases, including dense layouts, vertical stacking, nested composite assets, low resolution, and cluttered backgrounds. We compare TableVerse against three state-of-the-art single-view scene reconstruction baselines: \textbf{MIDI}~\citep{huang2025midi}, \textbf{SAM3D}~\citep{chen2025sam}, and \textbf{SceneMaker}~\citep{shi2025scenemaker}. The quantitative results are summarized in Table~\ref{tab:main_comparisons}.

\begin{figure}[htbp]
\centering
\includegraphics[width=1.0\linewidth]{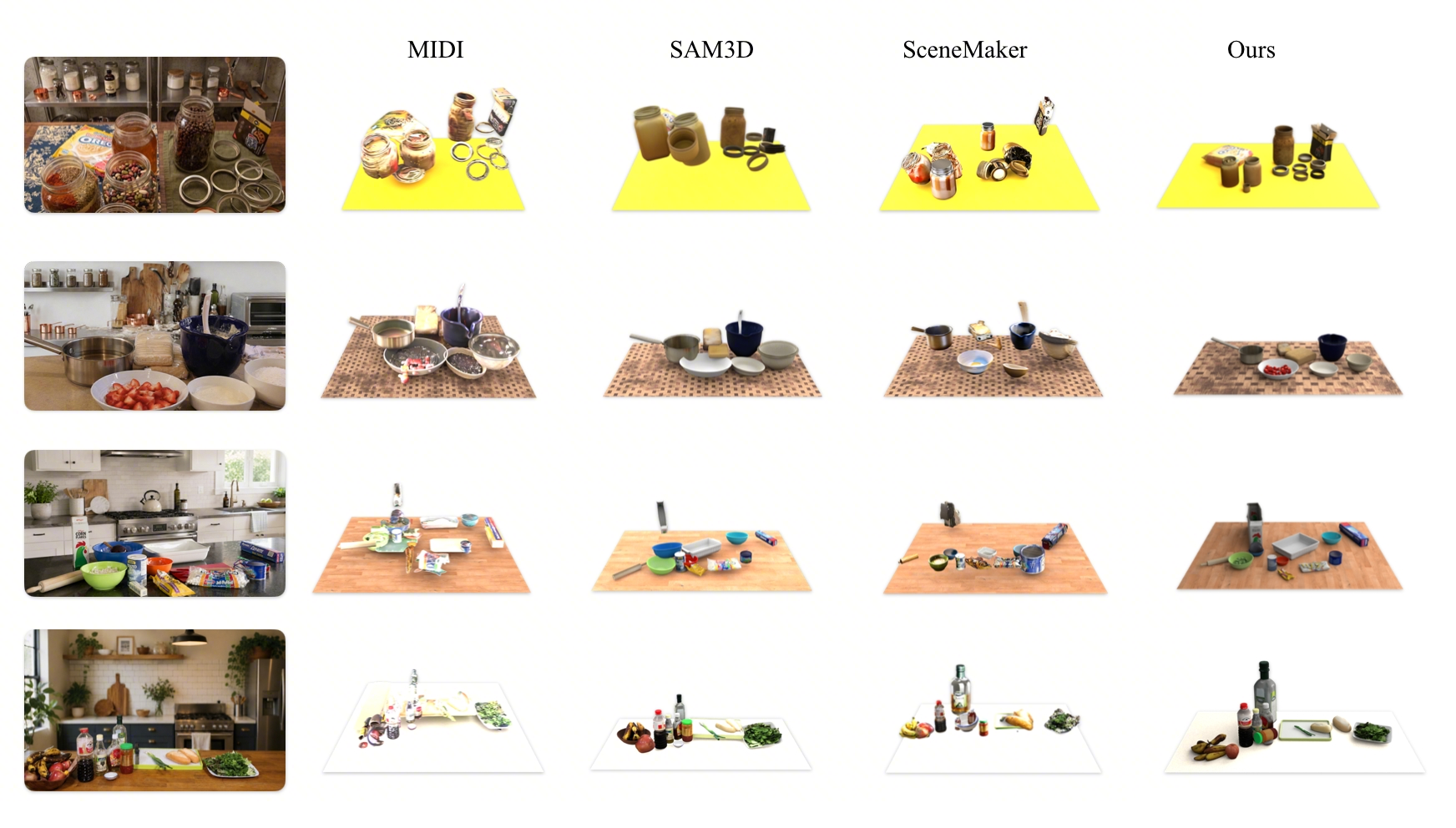}
\caption{Comparisons with Alternative Methods.}
\label{fig:Comparisons with Alternative Methods.}
\end{figure}

\begin{table}[htbp]
\centering
\caption{Quantitative comparison against baseline single-view scene synthesis methods across 100 in-the-wild test samples. Layout Fidelity (LF), Visual Quality (VQ), and Geometry Quality (GQ) are sub-dimensions of the MLLM-based GPT-Score.}
\label{tab:main_comparisons}
\small
\begin{tabular}{lccccc}
\toprule
\multirow{2}{*}{\textbf{Method}} & \multirow{2}{*}{\textbf{Scene Collision Rate (\%) $\downarrow$}} & \multicolumn{4}{c}{\textbf{GPT-Score}} \\ \cmidrule(lr){3-6} 
 &  & \textbf{LF} $\uparrow$ & \textbf{VQ} $\uparrow$ & \textbf{GQ} $\uparrow$ & \textbf{Avg. Rank} $\downarrow$ \\ \midrule
MIDI & 90.0\% & 5.81 & 5.08 & 4.83 & 2.73 \\
SAM3D & 81.0\% & 5.73 & 5.44 & 5.47 & 2.67 \\
SceneMaker & 72.0\% & 4.90 & 5.48 & 4.85 & 3.22 \\
\rowcolor[HTML]{EFEFEF} 
\textbf{TableVerse (Ours)} & \textbf{0.0\%} & \textbf{7.14} & \textbf{7.08} & \textbf{7.03} & \textbf{1.38} \\ \bottomrule
\end{tabular}
\end{table}

\paragraph{Quantitative and Qualitative Analysis}
As delineated in Table~\ref{tab:main_comparisons}, \textbf{TableVerse} significantly outperforms all baselines across all dimensions, notably securing an absolute \textbf{0.0\%} Scene Collision Rate alongside the top average cross-method ranking (\textbf{1.38}) within the multi-dimensional GPT-Score evaluation. Prior paradigms exhibit severe structural and physical vulnerabilities when handling unscripted internet data. Specifically, \textit{SceneMaker} degrades heavily under dense clutter and complex backgrounds, yielding a severely compromised layout fidelity (\textbf{4.90} LF) and the worst overall rank (\textbf{3.22}). Meanwhile, \textit{SAM3D} completely lacks rigid metric alignment and physical boundary constraints, triggering a catastrophic scene collision rate of \textbf{81.0\%} that renders the generated environments completely unsimulable. Similarly, \textit{MIDI} suffers from the most pervasive mesh interpenetrations (\textbf{90.0\%} collision rate) and fails completely to resolve fine-grained geometric structures, resulting in a highly deficient geometry score (\textbf{4.83} GQ). In sharp contrast, by combining deterministic feed-forward depth alignment with our layout-preserving LCCR optimization and physical stabilization, TableVerse entirely eliminates physical anomalies while establishing a massive lead in layout fidelity (\textbf{7.14} LF), visual quality (\textbf{7.08} VQ), and high-fidelity geometric recovery (\textbf{7.03} GQ) that faithfully matches real-world references.

\subsection{Ablation Study}
\begin{figure}[htbp]
\centering
\includegraphics[width=0.9\linewidth]{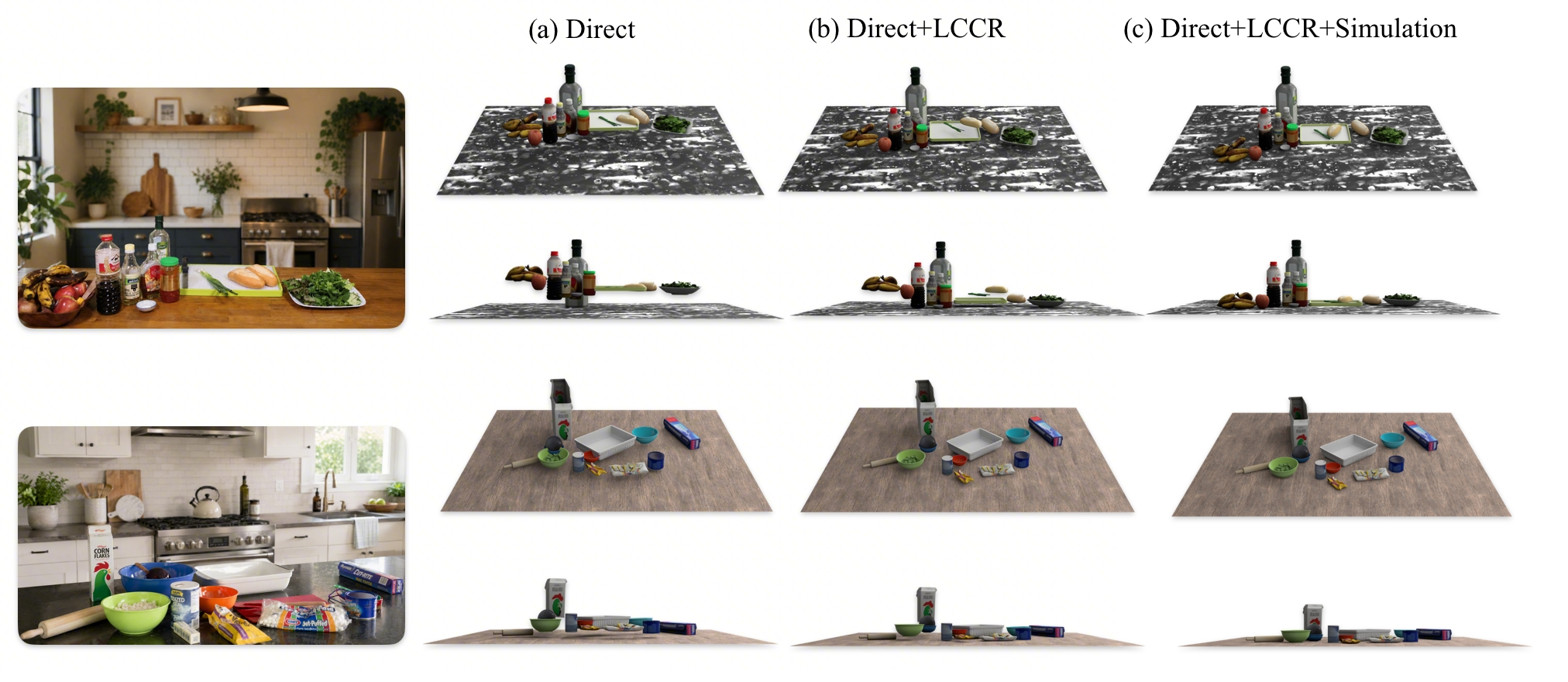}
\caption{Qualitative ablation of the LCCR module. (a) Direct alignment exhibits severe, unsimulable mesh interpenetrations. (b) Direct + LCCR eliminates collisions geometrically but introduces floating artifacts and micro-gaps. (c) Our full pipeline leverages MuJoCo simulation to settle objects into a stable, physics-ready resting state.}
\label{fig:lccr_visual_ablation}
\vspace{-5pt} 
\end{figure}

To verify the efficacy of the \textbf{Layout-Consistent Collision Rectification (LCCR)} module, we evaluate the 100 tabletop environments from our curated test set under three configurations: (1) \textbf{Direct}: baseline alignment of 3D meshes to feed-forward depth point clouds without collision handling; (2) \textbf{Direct + LCCR}: geometric mesh disentanglement via horizontal radial expansion and vertical footprint-based sorting; and (3) \textbf{Direct + LCCR + Simulation (Ours)}: our complete framework where rectified layouts undergo a forward MuJoCo simulation to reach a stable resting state. Quantitative and qualitative results are summarized in Table~\ref{tab:lccr_ablation} and Figure~\ref{fig:lccr_visual_ablation}, respectively.

\begin{table}[htbp]
\centering
\caption{Ablation study of the LCCR module across 100 evaluation scenes.}
\label{tab:lccr_ablation}
\begin{tabular}{lc}
\toprule
\textbf{Configuration} & \textbf{Scene Collision Rate (\%)} $\downarrow$ \\ \midrule
Direct & 79.0\% \\
Direct + LCCR & \textbf{0.0\%} \\
\rowcolor[HTML]{EFEFEF} \textbf{Direct + LCCR + Simulation (Ours)} & \textbf{0.0\%} \\ \bottomrule
\end{tabular}
\end{table}

\paragraph{Result Analysis}
As shown in Table~\ref{tab:lccr_ablation}, naive \textit{Direct} alignment yields a high \textbf{79.0\%} collision rate due to perception noise, making raw outputs unsimulable (Figure~\ref{fig:lccr_visual_ablation}a). Our \textit{LCCR} algorithm completely eliminates volumetric overlap to achieve a \textbf{0.0\%} collision rate (Figure~\ref{fig:lccr_visual_ablation}b), validating our layout-preserving optimization. While geometric rectification ensures zero collisions, pure rigid translations leave artificial micro-gaps or floating assets. The final forward \textit{Simulation} phase settles objects under gravity to close these remaining gaps (Figure~\ref{fig:lccr_visual_ablation}c), securing contact-validated, physics-ready digital twins.


\section{Limitation}
While our method demonstrates stable performance when processing internet data, it still has some limitations. These limitations stem from the constraints of SAM3D and the data input. Sometimes, objects within a container have low resolution and occupy fewer pixels, causing SAM3D to generate completely different objects. Additionally, generating 3D models for all objects in the entire scene using SAM3D is time-consuming. In future work, we will consider enabling the 3D generated models to perform one-time inference on the scene, thereby improving the speed of batch processing.
	

\section{Conclusion}
In this paper, we propose a fully automated, non-human-interventional desktop scene synthesis pipeline for Real2Sim, enabling the generation of simulation-ready scene assets from a single in-the-wild image. This pipeline incorporates a combined asset synthesis method, addressing the previous limitation of objects within containers being "visible but not tangible." Our proposed LCCR+Mujoco approach allows the generated scene assets to be directly loaded and used in the simulation engine. Furthermore, based on this automated processing pipeline, we generated a dataset, the TableVerse-100K, and used it to generate trajectory data in simulations. This dataset exhibits high asset and layout diversity, offering valuable insights for the field of robot manipulation.


\clearpage


\bibliographystyle{plainnat}
\bibliography{references}  

@inproceedings{huang2025midi,
  title     = {{MIDI}: Multi-Instance Diffusion for Single Image to {3D} Scene Generation},
  author    = {Huang, Zehuan and Guo, Yuan-Chen and An, Xingqiao and Yang, Yunhan and
               Li, Yangguang and Zou, Zi-Xin and Liang, Ding and Liu, Xihui and
               Cao, Yan-Pei and Sheng, Lu},
  booktitle = {Proceedings of the IEEE/CVF Conference on Computer Vision and Pattern Recognition (CVPR)},
  pages     = {23646--23657},
  year      = {2025}
}

@article{chen2025sam,
  title={Sam 3d: 3dfy anything in images},
  author={Chen, Xingyu and Chu, Fu-Jen and Gleize, Pierre and Liang, Kevin J and Sax, Alexander and Tang, Hao and Wang, Weiyao and Guo, Michelle and Hardin, Thibaut and Li, Xiang and others},
  journal={arXiv preprint arXiv:2511.16624},
  year={2025}
}

@article{wang2025tabletopgen,
  title={TabletopGen: Instance-Level Interactive 3D Tabletop Scene Generation from Text or Single Image},
  author={Wang, Ziqian and He, Yonghao and Yang, Licheng and Zou, Wei and Ma, Hongxuan and Liu, Liu and Sui, Wei and Guo, Yuxin and Su, Hu},
  journal={arXiv preprint arXiv:2512.01204},
  year={2025}
}

@article{ren2024grounded,
  title={Grounded sam: Assembling open-world models for diverse visual tasks},
  author={Ren, Tianhe and Liu, Shilong and Zeng, Ailing and Lin, Jing and Li, Kunchang and Cao, He and Chen, Jiayu and Huang, Xinyu and Chen, Yukang and Yan, Feng and others},
  journal={arXiv preprint arXiv:2401.14159},
  year={2024}
}

@article{lin2025depth,
  title={Depth anything 3: Recovering the visual space from any views},
  author={Lin, Haotong and Chen, Sili and Liew, Junhao and Chen, Donny Y and Li, Zhenyu and Shi, Guang and Feng, Jiashi and Kang, Bingyi},
  journal={arXiv preprint arXiv:2511.10647},
  year={2025}
}

@inproceedings{ravi2025sam,
  title={Sam 2: Segment anything in images and videos},
  author={Ravi, Nikhila and Gabeur, Valentin and Hu, Yuan-Ting and Hu, Ronghang and Ryali, Chaitanya and Ma, Tengyu and Khedr, Haitham and R{\"a}dle, Roman and Rolland, Chloe and Gustafson, Laura and others},
  booktitle={International Conference on Learning Representations},
  volume={2025},
  pages={28085--28128},
  year={2025}
}

@article{wei2022approximate,
  title={Approximate convex decomposition for 3d meshes with collision-aware concavity and tree search},
  author={Wei, Xinyue and Liu, Minghua and Ling, Zhan and Su, Hao},
  journal={ACM Transactions on Graphics (TOG)},
  volume={41},
  number={4},
  pages={1--18},
  year={2022},
  publisher={ACM New York, NY, USA}
}

@article{shi2025scenemaker,
  title={SceneMaker: Open-set 3D Scene Generation with Decoupled De-occlusion and Pose Estimation Model},
  author={Shi, Yukai and Li, Weiyu and Wang, Zihao and Li, Hongyang and Chen, Xingyu and Tan, Ping and Zhang, Lei},
  journal={arXiv preprint arXiv:2512.10957},
  year={2025}
}

@article{Yang2023HolodeckLG,
  title={Holodeck: Language Guided Generation of 3D Embodied AI Environments},
  author={Yue Yang and Fan-Yun Sun and Luca Weihs and Eli VanderBilt and Alvaro Herrasti and Winson Han and Jiajun Wu and Nick Haber and Ranjay Krishna and Lingjie Liu and Chris Callison-Burch and Mark Yatskar and Aniruddha Kembhavi and Christopher Clark},
  journal={2024 IEEE/CVF Conference on Computer Vision and Pattern Recognition (CVPR)},
  year={2023},
  pages={16277-16287},
  url={https://api.semanticscholar.org/CorpusID:266210109}
}

@article{Feng2023LayoutGPTCV,
  title={LayoutGPT: Compositional Visual Planning and Generation with Large Language Models},
  author={Weixi Feng and Wanrong Zhu and Tsu-Jui Fu and Varun Jampani and Arjun Reddy Akula and Xuehai He and Sugato Basu and Xin Eric Wang and William Yang Wang},
  journal={ArXiv},
  year={2023},
  volume={abs/2305.15393},
  url={https://api.semanticscholar.org/CorpusID:258865950}
}

@article{hao2026mesatask,
  title={Mesatask: Towards task-driven tabletop scene generation via 3d spatial reasoning},
  author={Hao, Jinkun and Liang, Naifu and Luo, Zhen and Xu, Xudong and Zhong, Weipeng and Yi, Ran and Jin, Yichen and Lyu, Zhaoyang and Zheng, Feng and Ma, Lizhuang and others},
  journal={Advances in neural information processing systems},
  volume={38},
  pages={122057--122099},
  year={2026}
}

@inproceedings{yu2025metascenes,
  title={Metascenes: Towards automated replica creation for real-world 3d scans},
  author={Yu, Huangyue and Jia, Baoxiong and Chen, Yixin and Yang, Yandan and Li, Puhao and Su, Rongpeng and Li, Jiaxin and Li, Qing and Liang, Wei and Zhu, Song-Chun and others},
  booktitle={Proceedings of the Computer Vision and Pattern Recognition Conference},
  pages={1667--1679},
  year={2025}
}

@article{Fu20203DFRONT3F,
  title={3D-FRONT: 3D Furnished Rooms with layOuts and semaNTics},
  author={Huan Fu and Bowen Cai and Lin Gao and Ling-Xiao Zhang and Cao Li and Zengqi Xun and Chengyue Sun and Yiyun Fei and Yu-qiong Zheng and Ying Li and Yi Liu and Peng Liu and Lin Ma and Le Weng and Xiaohang Hu and Xin Ma and Qian Qian and Rongfei Jia and Binqiang Zhao and Hao Helen Zhang},
  journal={2021 IEEE/CVF International Conference on Computer Vision (ICCV)},
  year={2020},
  pages={10913-10922},
  url={https://api.semanticscholar.org/CorpusID:227013144}
}

@article{Dai2017ScanNetR3,
  title={ScanNet: Richly-Annotated 3D Reconstructions of Indoor Scenes},
  author={Angela Dai and Angel X. Chang and Manolis Savva and Maciej Halber and Thomas A. Funkhouser and Matthias Nie{\ss}ner},
  journal={2017 IEEE Conference on Computer Vision and Pattern Recognition (CVPR)},
  year={2017},
  pages={2432-2443},
  url={https://api.semanticscholar.org/CorpusID:7684883}
}

@article{Zhou2025IL3DAL,
  title={IL3D: A Large-Scale Indoor Layout Dataset for LLM-Driven 3D Scene Generation},
  author={Wenxu Zhou and Kaixuan Nie and Hang Du and Dong Yin and Wei Huang and Siqi Guo and Xiaobo Zhang and Peng Hu},
  journal={ArXiv},
  year={2025},
  volume={abs/2510.12095},
  url={https://api.semanticscholar.org/CorpusID:282064609}
}

@article{Xu2022TOSceneAL,
  title={TO-Scene: A Large-scale Dataset for Understanding 3D Tabletop Scenes},
  author={Mutian Xu and P. Chen and Haolin Liu and Xiaoguang Han},
  journal={ArXiv},
  year={2022},
  volume={abs/2203.09440},
  url={https://api.semanticscholar.org/CorpusID:247519171}
}

@article{Todorov2012MuJoCoAP,
  title={MuJoCo: A physics engine for model-based control},
  author={Emanuel Todorov and Tom Erez and Yuval Tassa},
  journal={2012 IEEE/RSJ International Conference on Intelligent Robots and Systems},
  year={2012},
  pages={5026-5033},
  url={https://api.semanticscholar.org/CorpusID:5230692}
}

@article{google2025gemini25,
  title         = {Gemini 2.5: Pushing the Frontier with Advanced Reasoning, Multimodality, Long Context, and Next Generation Agentic Capabilities},
  author        = {{Gemini Team, Google}},
  year          = {2025},
  eprint        = {2507.06261},
  archivePrefix = {arXiv},
  primaryClass  = {cs.CL},
  doi           = {10.48550/arXiv.2507.06261}
}

@inproceedings{Seed2026Seed18MC,
  title={Seed1.8 Model Card: Towards Generalized Real-World Agency},
  author={ByteDance Seed},
  year={2026},
  url={https://api.semanticscholar.org/CorpusID:286762238}
}

@article{mahler2017dex,
  title={Dex-net 2.0: Deep learning to plan robust grasps with synthetic point clouds and analytic grasp metrics},
  author={Mahler, Jeffrey and Liang, Jacky and Niyaz, Sherdil and Laskey, Michael and Doan, Richard and Liu, Xinyu and Ojea, Juan Aparicio and Goldberg, Ken},
  journal={arXiv preprint arXiv:1703.09312},
  year={2017}
}

@article{ten2017grasp,
  title={Grasp pose detection in point clouds},
  author={Ten Pas, Andreas and Gualtieri, Marcus and Saenko, Kate and Platt, Robert},
  journal={The International Journal of Robotics Research},
  volume={36},
  number={13-14},
  pages={1455--1473},
  year={2017},
  publisher={SAGE Publications Sage UK: London, England}
}

@inproceedings{liang2019pointnetgpd,
  title={Pointnetgpd: Detecting grasp configurations from point sets},
  author={Liang, Hongzhuo and Ma, Xiaojian and Li, Shuang and G{\"o}rner, Michael and Tang, Song and Fang, Bin and Sun, Fuchun and Zhang, Jianwei},
  booktitle={2019 international conference on robotics and automation (ICRA)},
  pages={3629--3635},
  year={2019},
  organization={IEEE}
}

@inproceedings{sundermeyer2021contact,
  title={Contact-graspnet: Efficient 6-dof grasp generation in cluttered scenes},
  author={Sundermeyer, Martin and Mousavian, Arsalan and Triebel, Rudolph and Fox, Dieter},
  booktitle={2021 IEEE international conference on robotics and automation (ICRA)},
  pages={13438--13444},
  year={2021},
  organization={IEEE}
}

@inproceedings{fang2020graspnet,
  title={Graspnet-1billion: A large-scale benchmark for general object grasping},
  author={Fang, Hao-Shu and Wang, Chenxi and Gou, Minghao and Lu, Cewu},
  booktitle={Proceedings of the IEEE/CVF conference on computer vision and pattern recognition},
  pages={11444--11453},
  year={2020}
}

@article{murali2025graspgen,
  title={Graspgen: A diffusion-based framework for 6-dof grasping with on-generator training},
  author={Murali, Adithyavairavan and Sundaralingam, Balakumar and Chao, Yu-Wei and Yuan, Wentao and Yamada, Jun and Carlson, Mark and Ramos, Fabio and Birchfield, Stan and Fox, Dieter and Eppner, Clemens},
  journal={arXiv preprint arXiv:2507.13097},
  year={2025}
}

@inproceedings{schulman2013finding,
  title={Finding locally optimal, collision-free trajectories with sequential convex optimization.},
  author={Schulman, John and Ho, Jonathan and Lee, Alex X and Awwal, Ibrahim and Bradlow, Henry and Abbeel, Pieter},
  booktitle={Robotics: science and systems},
  volume={9},
  number={1},
  pages={1--10},
  year={2013},
  organization={Berlin, Germany}
}

@article{sundaralingam2023curobo,
  title={curobo: Parallelized collision-free minimum-jerk robot motion generation},
  author={Sundaralingam, Balakumar and Hari, Siva Kumar Sastry and Fishman, Adam and Garrett, Caelan and Van Wyk, Karl and Blukis, Valts and Millane, Alexander and Oleynikova, Helen and Handa, Ankur and Ramos, Fabio and others},
  journal={arXiv preprint arXiv:2310.17274},
  year={2023}
}

@inproceedings{liang2023code,
  title={Code as policies: Language model programs for embodied control},
  author={Liang, Jacky and Huang, Wenlong and Xia, Fei and Xu, Peng and Hausman, Karol and Ichter, Brian and Florence, Pete and Zeng, Andy},
  booktitle={2023 IEEE International conference on robotics and automation (ICRA)},
  pages={9493--9500},
  year={2023},
  organization={IEEE}
}

@article{huang2023voxposer,
  title={Voxposer: Composable 3d value maps for robotic manipulation with language models},
  author={Huang, Wenlong and Wang, Chen and Zhang, Ruohan and Li, Yunzhu and Wu, Jiajun and Fei-Fei, Li},
  journal={arXiv preprint arXiv:2307.05973},
  year={2023}
}

@inproceedings{gao2025genmanip,
  title={Genmanip: Llm-driven simulation for generalizable instruction-following manipulation},
  author={Gao, Ning and Chen, Yilun and Yang, Shuai and Chen, Xinyi and Tian, Yang and Li, Hao and Huang, Haifeng and Wang, Hanqing and Wang, Tai and Pang, Jiangmiao},
  booktitle={Proceedings of the Computer Vision and Pattern Recognition Conference},
  pages={12187--12198},
  year={2025}
}

\clearpage
\appendix
\raggedbottom  

\section{Open-Vocabulary Object Detection and Prompt Details}
\label{app:seed_detection}

To anchor unscripted, \textit{in-the-wild} internet images into simulation-ready tabletop layouts, our pipeline begins with an open-world visual perception front-end. We employ \textbf{Seed-1.8}~\citep{Seed2026Seed18MC} as our core open-vocabulary object detector to localize arbitrary, non-predefined target objects on the tabletop. Given a single-view visual input, the model takes a structured text prompt as a linguistic query to reason about the scene, filter out background clutter (e.g., human hands, body parts), and output precise 2D bounding boxes along with semantic category labels for each valid manipulable asset.

\paragraph{Seed-1.8 Hybrid Detection Prompt}
To guide the open-vocabulary detection process deterministically, the exact, compiled textual prompt template fed into Seed-1.8 is presented below. Note that this block automatically breaks across pages to accommodate the detailed architectural constraints:%

\begin{tcolorbox}[
    enhanced,
    breakable,
    before skip=0.8ex,         
    after skip=0.8ex,          
    colback=blue!3,           
    colframe=blue!60!black,    
    title={System Prompt for Hybrid Object Detection \& Scene Parsing}, 
    fonttitle=\bfseries\small,
    coltitle=white,
    attach boxed title to top left={yshift=-2mm, xshift=2mm},
    boxed title style={colback=blue!60!black, sharp corners=south}, 
    fontupper=\small,     
    arc=1mm
]
\noindent You are doing hybrid object detection. Produce \textbf{TWO} categories: (1) regular detections and (2) composite objects (\texttt{container\_fill}).

\vspace{0.5em}
\textbf{[Rules for All Objects]}
\begin{itemize}[leftmargin=1.5em, topsep=0.2ex, itemsep=0.3ex]
    \item \textbf{Exclude Human Elements:} Exclude the person and any hands/arms from detections and composites (do not output bounding boxes or labels like \texttt{'hand'}, \texttt{'person'}, \texttt{'arm'}).
    \item \textbf{Front-of-Person Heuristic:} If a full person (head + torso) is visible, first estimate the main person's bounding box. Only keep objects that are in front of the person (held in hands OR clearly not occluded by the person's body). Do \textbf{NOT} include objects behind the person (i.e., drop any object whose box has $>50\%$ area overlap with the person's box and is not held). If only a hand/arm is visible, treat it as no person and detect foreground normally.
    \item \textbf{Tabletop Surface:} Always detect the visible tabletop/supporting surface (\texttt{table}, \texttt{desk}, \texttt{countertop}, \texttt{work surface}). Return \textbf{exactly one} box with the label \texttt{'table'}. This box must also follow the front-of-person rule.
\end{itemize}

\vspace{0.5em}
\textbf{[Logic for Containers]}
\begin{itemize}[leftmargin=1.5em, topsep=0.2ex, itemsep=0.3ex]
    \item \textbf{Content Constraints:} For \texttt{container\_fill} contents, only include discrete, solid, standalone objects with clear boundaries that can be generated as independent 3D assets. Do \textbf{NOT} include liquids (water, coffee, etc.), powders/granules, sauces/foam, reflections/glare, shadows, or generic materials.
    \item \textbf{Fallback Rule:} If a container is empty, or contains \textbf{ANY} non-solid/non-3D-generatable content (like liquid/sauce), do \textbf{NOT} create a composite for it. Instead, you \textbf{MUST} include that container as a single regular object in the \texttt{'detections'} list. Never ignore a visible foreground container.
\end{itemize}

\vspace{0.5em}
\textbf{[BBox Representation \& Precision]}
\begin{itemize}[leftmargin=1.5em, topsep=0.2ex, itemsep=0.3ex]
    \item \textbf{No Grouping (Anti-Grouping):} For contents with \texttt{count > 1}, output \textbf{EXACTLY ONE} bbox that tightly covers \textbf{ONLY ONE} representative individual instance. Do \textbf{NOT} create a single large bbox that encompasses multiple instances or a pile. Pick the most visible one to draw the box, and use \texttt{'count'} to reflect the total number. Bbox coordinates $[xmin, ymin, xmax, ymax]$ must be normalized in $[0, 1000]$.
    \item \textbf{Completeness \& De-duplication:} Every foreground object must appear \textbf{EXACTLY once} in the output:
    \begin{enumerate}[label=\roman*., leftmargin=1.5em, noitemsep]
        \item Container with valid 3D-generatable contents $\rightarrow$ Output \textbf{ONLY} in \texttt{'composites'}.
        \item Container empty or with liquid/powder contents $\rightarrow$ Output \textbf{ONLY} in \texttt{'detections'}.
        \item Regular solid objects (not containers) $\rightarrow$ Output \textbf{ONLY} in \texttt{'detections'}.
    \end{enumerate}
\end{itemize}

\vspace{0.5em}
\textbf{[Metrics \& Descriptions]} \\
For \texttt{container\_fill}, provide faithful English descriptions for \texttt{container.description} and each \texttt{contents[i].description}. Include visible attributes (color, relative size, shape, material, texture, markings); do not speculate.

\vspace{0.5em}
\textbf{[Output JSON Schema]} \\
Output JSON only (no Markdown, no code fences). Follow this exact schema:
\begin{minipage}{\linewidth}
\begin{verbatim}
{
  "detections": [
    {
      "label": string,
      "bbox_xyxy": [xmin, ymin, xmax, ymax],
      "score": float,
      "description": string (optional)
    }
  ],
  "composites": [
    {
      "type": "container_fill",
      "description": string,
      "container": {
        "label": string,
        "bbox": [xmin, ymin, xmax, ymax],
        "description": string,
      },
      "contents": [
        {
          "label": string,
          "bbox": [xmin, ymin, xmax, ymax],
          "count": int,
          "description": string,
        }
      ]
    }
  ]
}
\end{verbatim}
\end{minipage}
\end{tcolorbox}

\paragraph{Post-Processing and Mask Seeding}
The raw 2D bounding boxes and JSON attributes generated by Seed-1.8 are structured to seamlessly bootstrap downstream modules. Labels falling into the \texttt{detections} category are passed directly to SAM2 for standard single-instance mask generation. For entities classified under \texttt{composites}, the outermost container coordinates serve as the macro-anchor, while individual bounding boxes in the \texttt{contents} array are expanded dynamically into pixel-level seeds. This hybrid topological definition directly feeds the hierarchical mesh-to-point-cloud grounding and layout consistency modules detailed in the main text.

\section{Detailed Dataset Curation and MLLM Annotation Pipeline}
\label{app:dataset_details}

In this section, we provide the underlying technological assets and prompt sources for the automated dataset curation, multi-view prompt execution, and multi-dimensional semantic annotation protocol driven by Gemini 2.5 Pro~\citep{google2025gemini25}. Following the macro-workflow described in Section 3.3 of the main text, the comprehensive and unedited system prompt used to curate the \textbf{TableVerse-100K Dataset} is provided in Box~\ref{box:dataset_curation_prompt}.

\begin{tcolorbox}[
    enhanced,
    breakable,            
    colback=blue!3,       
    colframe=blue!60!black, 
    title={System Prompt for Dataset Curation, Attribute Prediction \& Pick-and-Place Generation}, 
    fonttitle=\bfseries\small,
    coltitle=white,
    attach boxed title to top left={yshift=-2mm, xshift=2mm},
    boxed title style={colback=blue!60!black, sharp corners=south},
    fontupper=\small,     
    arc=1mm,
    label={box:dataset_curation_prompt}
]
\textbf{[Role and Context]} \\
You are an expert robotic vision and manipulation analysis system. You are presented with a tabletop manipulation scene captured from two distinct viewpoints:
\begin{itemize}[leftmargin=1.5em, noitemsep]
    \item \textbf{LEFT:} a front view captured from a slightly elevated (tilted) camera angle.
    \item \textbf{RIGHT:} a top-down view of the same scene.
\end{itemize}
Each object in the scene is annotated with a 2D bounding box, a numeric ID, and a text label provided separately by the user in the form: \texttt{\{id: label\}}. Read the label mapping carefully before judging.

\vspace{0.5em}
\textbf{[Your Task]} \\
Produce \textbf{SEVEN} outputs for the scene: (A) USABILITY GATE, (B) SEVERE LABEL ERROR CHECK, (C) LABEL CONFIDENCE, (D) PHYSICAL PROPERTIES, (E) QUALITY SCORE, (F) OBJECT RELABELING, and (G) PICK \& PLACE TASK GENERATION. Parts A--E guide data curation. Parts F and G \textbf{MUST} be generated regardless of whether the scene is usable or not.

\vspace{0.5em}
\textbf{[PART A. Usability Gate]} \\
Set \texttt{usable = false} if \textbf{ANY} of the following hard-failure conditions is met (otherwise \texttt{usable = true}):
\begin{itemize}[leftmargin=1.5em, topsep=0.2ex, itemsep=0.3ex]
    \item \texttt{EXTREME\_REGIONAL\_PARTITION}: On the \textbf{TOP} view, objects form two or more spatially separated clusters such that the empty band between clusters is wider than roughly the longest dimension of the larger cluster, OR objects are isolated far from the main group such that a single robot workspace cannot reasonably cover them.
    \item \texttt{NON\_TABLETOP\_OBJECT}: The scene contains objects that should not realistically appear on a tabletop in an everyday manipulation scene, such as large home appliances, humans/body parts, or live animals. List offending IDs in \texttt{evidence\_ids.NON\_TABLETOP\_OBJECT}.
\end{itemize}
If \texttt{usable = false}, cap the final quality score at 4.

\vspace{0.5em}
\textbf{[PART B. Severe Label Error Check]} \\
Determine whether the scene contains any text label that is clearly and obviously wrong (e.g., box labeled ``scissors'' shows a cup). Apply a \textbf{HIGH} bar. Do \textbf{NOT} count as severe errors: synonyms (\textit{mug} vs \textit{cup}), hypernyms (\textit{tool} for screwdriver), stylistic/material inaccuracies, low-poly geometry, or generic labels (\textit{object}). If errors exist, set \texttt{has\_severe\_label\_error = true} and provide corrected labels.

\vspace{0.5em}
\textbf{[PART C \& D. Label Confidence \& Physical Properties]}
\begin{itemize}[leftmargin=1.5em, topsep=0.2ex, itemsep=0.3ex]
    \item \textbf{Label Confidence:} Output a confidence value in $[0, 1]$ for \textbf{EVERY} object ID. Higher values mean higher certainty. If confidence $< 0.7$, estimate physical properties conservatively using the provided label name only.
    \item \textbf{Physical Properties:} Predict \texttt{mass\_kg} (positive real number based on apparent scale and material), \texttt{has\_hinge} (boolean), and \texttt{hinge\_category} (classify strictly into allowed enums like \textit{Bottle, Box, Bucket, Laptop, Pliers, Scissors, TrashCan} or \textit{null}).
\end{itemize}

\vspace{0.5em}
\textbf{[PART E. Quality Score (1--10)]} \\
Rate \textbf{OVERALL QUALITY} on an integer scale 1--10 based on four weighted criteria:
\begin{itemize}[leftmargin=1.5em, topsep=0.2ex, itemsep=0.3ex]
    \item \textbf{Object Count \& Category Diversity (45\%):} Reward scenes with clutter density $n \ge 5$. Apply a strict penalty if the object count drops below 5.
    \item \textbf{Geometric Plausibility (15\%):} Penalize distorted, broken, melted, or non-physical geometry.
    \item \textbf{Graspability for Robots (15\%):} Favor objects with clear graspable features; penalize objects too large, heavy, or flat to grasp.
    \item \textbf{Layout Clarity \& Reasonableness (25\%):} Reward clean multi-object setups; penalize severe non-physical clipping, floating objects, or rendering glitches.
\end{itemize}

\vspace{0.5em}
\textbf{[PART F. Object Relabeling]} \\
For \textbf{EVERY} annotated object ID, independently assign a concise \texttt{relabel\_name} from visual evidence alone, and a list of \texttt{distinguishing\_phrases} containing multiple short noun phrases with adjectives (e.g., \textit{``the ceramic cup nearest to the camera''}) to isolate identical instances. Never use artificial render suffixes like \texttt{bowl\_1}.

\vspace{0.5em}
\textbf{[PART G. Pick \& Place Task Generation]} \\
Generate a list of pick-and-place tasks for this scene regardless of usability.
\begin{itemize}[leftmargin=1.5em, topsep=0.2ex, itemsep=0.3ex]
    \item \textbf{Template:} Each description must exactly match: \texttt{Pick up <X> and place it <relation> <Y>}, where \texttt{<X>} is the natural-language name of the object to be picked (optionally with a disambiguating descriptor like color/position), \texttt{<Y>} is the reference/container object, and \texttt{<relation>} must be selected strictly from the fixed enum: \texttt{\{"on", "in", "into", "next to", "to the left of", "to the right of", "behind", "in front of", "inside", "outside"\}}.
    \item \textbf{Hard Constraints:} (1) \texttt{X\_id} and \texttt{Y\_id} must be different valid IDs. (2) Initial state must not satisfy the relation (judge jointly from top/front views). (3) Task must be physically feasible (X is graspable, target placement fits; use \textit{in/into/inside} only for genuine containers, and \textit{on} only for stable surfaces). (4) Daily-life plausibility must be preserved (avoid nonsensical pairings). (5) Diversity: generate as many tasks as naturally supported, cover different assets without repeating triples, and try to use at least 3 distinct relations across the list. (6) Return an empty list if no valid task can be constructed.
\end{itemize}

\vspace{0.5em}
\textbf{[Output JSON Schema]} \\
Return \textbf{ONLY} a valid JSON object without markdown fences or extra introductory/concluding prose:

\footnotesize 
\begin{verbatim}
{
  "usability": {
    "usable": boolean,
    "reason_codes": [
      "EXTREME_REGIONAL_PARTITION" | "NON_TABLETOP_OBJECT"
    ],
    "evidence_ids": { 
      "NON_TABLETOP_OBJECT": [int, ...] 
    }
  },
  "severe_label_error": {
    "has_severe_label_error": boolean,
    "evidence_ids": [int, ...],
    "correct_labels": { "string_id": "string_corrected_label" }
  },
  "label_confidence": { "string_id": float },
  "physical_properties": {
    "string_id": { 
      "mass_kg": float, 
      "has_hinge": boolean, 
      "hinge_category": string | null 
    }
  },
  "per_criterion": {
    "category_diversity": { "score": int },
    "geometric_plausibility": { 
      "score": int, 
      "bad_geometry_ids": [int, ...] 
    },
    "graspability": { 
      "score": int, 
      "graspable_ids": [int, ...], 
      "ungraspable_ids": [int, ...] 
    },
    "layout_clarity": { "score": int }
  },
  "object_relabeling": {
    "string_id": { 
      "relabel_name": string, 
      "distinguishing_phrases": [string, ...] 
    }
  },
  "final_score": int,
  "generated_tasks": [
    {
      "instruction": string,
      "X_id": int,
      "X_name": string,
      "Y_id": int,
      "Y_name": string,
      "relation": "on" | "in" | "into" | "next to" | "to the left of" 
                 | "to the right of" | "behind" | "in front of" 
                 | "inside" | "outside",
      "initial_relation_satisfied": false
    }
  ]
}
\end{verbatim}
\end{tcolorbox}

\section{Implementation and Evaluation Details}

\subsection{Baseline Implementation Details}
To ensure a rigorous and fair comparison, we standardize the visual input front-end across all evaluated baselines. Since \textbf{MIDI}~\citep{huang2025midi}, \textbf{SAM3D}~\citep{chen2025sam}, and \textbf{SceneMaker}~\citep{shi2025scenemaker} lack native open-vocabulary object discovery and autonomous mask extraction capabilities, we uniformly supply them with the high-fidelity instance masks generated by our perception pipeline. This isolate-and-compare design ensures that downstream performance discrepancies purely reflect each method's core 3D synthesis and spatial alignment capabilities.

\paragraph{SceneMaker Point Cloud Up-sampling}
SceneMaker~\citep{shi2025scenemaker} inherently conditions its pose and scale regression networks on a fixed $1024$-dimensional point cloud extracted per instance. However, our \textit{in-the-wild} evaluation benchmark presents severe real-world challenges, including low visual resolution and dense object clustering. Under these unstructured conditions, heavily occluded or small objects yield minuscule mask pixel footprints, resulting in deficient and sparse point counts that trigger catastrophic failures in 6-DoF pose estimation. To mitigate this data-sparsity artifact, prevent object omission, and safeguard regression stability, we implement an instance-level point cloud up-sampling step. For any instance mask generating fewer than 1024 points, we perform nearest-neighbor up-sampling to bring the coordinate set to exactly 1024 points before forwarding it to SceneMaker's conditioning network.

\subsection{GPT-Score Evaluation Prompt Source}
The exact evaluation prompt deployed to instruct the Multimodal Large Language Model (MLLM) for calculating the multi-dimensional GPT-Score (comprising Layout Fidelity, Visual Quality, and Geometry Quality) is detailed below:

\begin{figure}[htbp]
    \centering
    \includegraphics[width=0.95\linewidth]{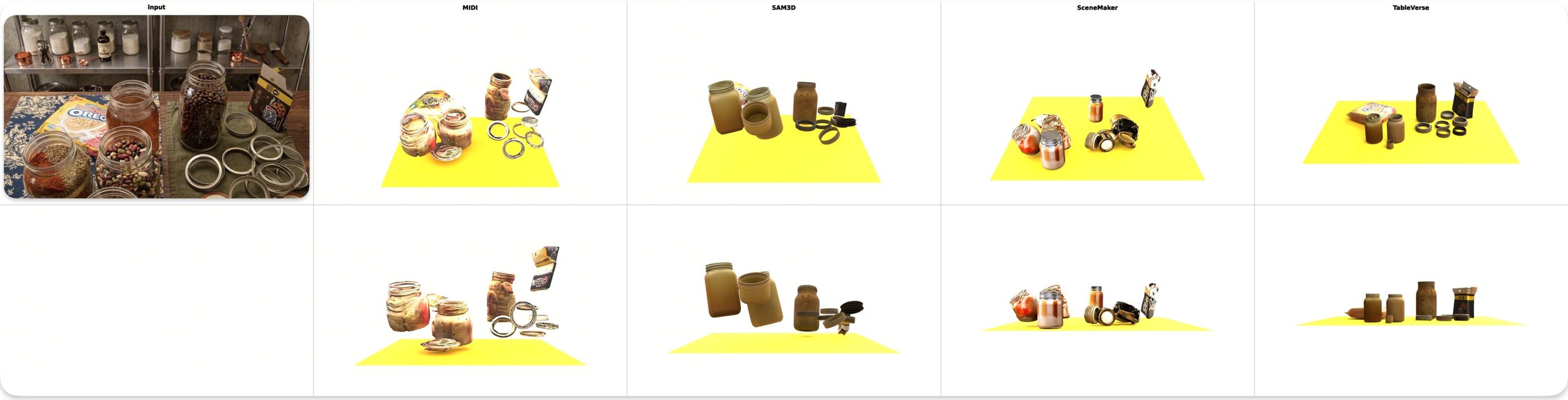}
    \caption{An example of the concatenated 2$\times$5 grid image provided as the input observation to the MLLM. Row 1 contains the real-world reference image followed by front-oblique renderings of the four methods. Row 2 consists of a blank slot followed by pure front-view renderings used to inspect vertical alignment and physical artifacts.}
    \label{fig:gpt_prompt_input}
\end{figure}

\begin{tcolorbox}[
    enhanced,
    breakable,
    colback=blue!3,       
    colframe=blue!60!black,    
    title={System Prompt for Real-to-Sim Evaluation}, 
    fonttitle=\bfseries\small,
    coltitle=white,
    attach boxed title to top left={yshift=-2mm, xshift=2mm},
    boxed title style={colback=blue!60!black, sharp corners=south},
    fontupper=\small\ttfamily, 
    arc=1mm
]
You are an expert evaluator for tabletop real-to-sim scene reconstruction.

\vspace{0.5em}
\textbf{[Input Stitched Image Layout]} \\
The input is a grid of 2 rows and 5 columns:
\begin{itemize}[leftmargin=1.5em, itemsep=0.1em]
    \item \textbf{Row 1, Col 1:} Input observation image (Reference).
    \item \textbf{Row 1, Col 2--5:} Front-oblique renderings (slight top-down angle) for methods in order: Column 2: MIDI | Column 3: SAM3D | Column 4: SceneMaker | Column 5: TableVerse.
    \item \textbf{Row 2, Col 1:} Blank.
    \item \textbf{Row 2, Col 2--5:} Pure front-view renderings (NO downward angle) for the same methods in the exact same order.
\end{itemize}

\vspace{0.5em}
\textbf{[Evaluation Scope]} \\
- Evaluate ONLY the tabletop scene; ignore the background.
- CRITICAL: Reconstructed 3D scenes contain ONLY objects, NOT the tabletop surface itself.
- The reference image may be captured from non-standard viewpoints. Reconstructed front views must use the default upward direction as the normal direction pointing vertically away from the tabletop.

\vspace{0.5em}
\textbf{[Evaluation Metrics] (Score: 1--10)}
\begin{enumerate}[leftmargin=1.5em, itemsep=0.3em]
    \item \textbf{Layout Fidelity (LF):} Measures preservation of the reference layout. Focus on object-to-object spatial relationships, scale, orientation (pose), and position. Penalize missing/extra objects or wrong poses.
    \item \textbf{Visual Quality (VQ):} Measures rendering and texture naturalness. Reward clear, recognizable, natural-looking assets. Penalize deformed shapes, poor textures, or severe artifacts.
    \item \textbf{Geometry Quality (GQ):} Measures if the object geometry matches the reference. Reward correct category, shape proportions, and structures. Penalize broken, melted, or collapsed assets.
\end{enumerate}

\vspace{0.5em}
\textbf{[Overall Ranking Rules]} \\
Rank 1 (best) to Rank 4 (worst). Ranks must be unique integers (1, 2, 3, 4). Base it on LF, VQ, GQ, and overall simulation readiness.

\vspace{0.5em}
\textbf{[Output Format]} \\
Return ONLY a valid JSON object without markdown code blocks or extra prose. Ensure "reason" strings are single-line with no raw newlines.
\begin{center}
[Exact JSON Schema as defined in the source code]
\end{center}
\end{tcolorbox}

\section{Algorithmic and Mathematical Details of Trajectory Generation}
\label{app:trajectory_algorithmic_details}
\subsection{Point Cloud Pre-processing and Multi-Stage Grasp Relaxation}
Before feeding the target object's surface vertices into the grasp detection network, the raw extracted point cloud underwent an initial alignment to minimize translation invariance biases during deep inference. Let $\mathcal{P}_{\text{raw}} = \{\mathbf{p}_i\}_{i=1}^N \in \mathbb{R}^{N \times 3}$ denote the initially sampled point cloud on the asset's surface. We compute the spatial centroid of the object as:
\begin{equation}
\mathbf{t}_{\text{center}} = \frac{1}{N} \sum_{i=1}^N \mathbf{p}_i
\end{equation}
The point cloud is then rigidly translated to the canonical origin, yielding $\mathcal{P} = \{\mathbf{p}_i - \mathbf{t}_{\text{center}}\}$, which is subsequently ingested by GraspGen to predict the set of 6D candidates $\mathcal{G}$.

To guarantee execution robustness in heavily occluded or dense environments where the strict top-down constraint ($\vec{a}_i \cdot \vec{z}_w \le \gamma_{\text{strict}}$) yields an empty set, the framework falls back onto a hierarchical, multi-stage relaxation arbitration loop:
\begin{itemize}
    \item \textbf{Stage 1 (Strict Top-Down):} Filters candidates whose approach vectors are strictly aligned within a narrow downward cone defined by $\gamma_{\text{strict}}$.
    \item \textbf{Stage 2 (Hemispheric Downward):} If Stage 1 returns zero valid poses, the directional boundary is adaptively relaxed to accept any generalized downward-facing orientation satisfying a hemispheric check:
    \begin{equation}
    \vec{a}_i \cdot \vec{z}_w < 0
    \end{equation}
    \item \textbf{Stage 3 (Confidence Fallback):} If the candidate set remains completely unpopulated due to geometry constraints, the directional filtering is bypassed entirely. The system dynamically falls back to the absolute network confidence, selecting the optimal grasp $g^*$ via:
    \begin{equation}
    g^* = \arg\max_{g_i \in \mathcal{G}} s_i
    \end{equation}
    where $s_i$ denotes the scoring scalar outputted by the network decoder for candidate $g_i$.
\end{itemize}

\subsection{Mathematical Construction of Relation-Specific Placement Domains}
The main text denotes $\Omega_{\mathcal{R}}(\mathcal{O}_{\text{ref}})$ as the valid localized bounding region for sampling placement coordinates. Depending on the semantic spatial predicate $\mathcal{R}$, this sampling space is analytically constructed from the 3D bounding box half-extents $(b_x, b_y, b_z)$ and centroid $(c_x, c_y, c_z)$ of the reference asset $\mathcal{O}_{\text{ref}}$:
\begin{itemize}
    \item \textbf{Containment and Stacking ($\mathcal{R} \in \{\text{top}, \text{in}\}$):} For stacking on flat surfaces, the placement domain is mathematically restricted to a 2D horizontal plane bounded by the reference asset's upper face:
    \begin{equation}
    \Omega_{\text{top}} = \left\{ (x, y, z) \mid x \in [c_x - b_x, c_x + b_x], \, y \in [c_y - b_y, c_y + b_y], \, z = c_z + b_z \right\}
    \end{equation}
    \item \textbf{Directional Adjacency ($\mathcal{R} \in \{\text{left}, \text{right}, \text{front}, \text{back}\}$):} For directional arrangements, the domain is formulated as a projected half-space offset along the principal coordinate axes of $\mathcal{O}_{\text{ref}}$. For instance, the region $\Omega_{\text{right}}$ along the world $Y$-axis is defined as:
    \begin{equation}
    \Omega_{\text{right}} = \left\{ (x, y, z) \mid x \in [c_x - b_x, c_x + b_x], \, y > c_y + b_y + \delta_{\text{margin}}, \, z = c_z - b_z \right\}
    \end{equation}
    where $\delta_{\text{margin}}$ represents a baseline physical parsing buffer to prevent initial contact.
\end{itemize}
Candidates are iteratively sampled within these defined boundary sets using a rejection sampling loop until the AABB non-penetration criteria formalized in the main text are satisfied.

\subsection{Coordinate Alignment and Hybrid Obstacle Arbitration}
Prior to generating joint-space trajectories via cuRobo, all non-target scene items are dynamically mapped into the local frame of the robot base $\mathcal{F}_{\text{robot}}$. Let $\mathbf{T}_{\text{world}\to\text{robot}} \in SE(3)$ denote the known kinematic transform of the manipulator base relative to the world coordinate system. For any scene asset residing at $\mathbf{T}_{\text{asset}} \in SE(3)$ in the world frame, its planning coordinate is evaluated as:
\begin{equation}
\mathbf{T}_{\text{planning}} = \mathbf{T}_{\text{world}\to\text{robot}} \cdot \mathbf{T}_{\text{asset}}
\end{equation}

To maximize motion optimization success in narrow spaces (such as the interior of hollow containers or clustered shelves), the planner operates a hybrid collision-representation hierarchy. Environmental boundaries are preferentially extracted as high-fidelity non-convex triangular meshes ($\mathcal{M}$) to accurately match real-world geometries. If the mesh generation pipeline encounters a non-manifold vertex configuration or topological degeneracy, the planning scene dynamically activates a fallback arbitrator, converting that specific asset into a conservative, circumscribed Axis-Aligned Bounding Box volume. This mathematical fallback prevents optimization divergence and preserves the algorithmic continuity of the trajectory generation loop.

\end{document}